\documentclass[11pt]{article}

\usepackage[final]{acl}
\usepackage{salute}

\usepackage{times}
\usepackage{latexsym}

\usepackage[T1]{fontenc}

\usepackage[utf8]{inputenc}

\usepackage{microtype}

\usepackage{inconsolata}

\usepackage{graphicx}

\title{SALUTE: Benchmarking and Adapting LLMs for the Defense Domain}

\author{
\textbf{Hyeongcheol Park\textsuperscript{1} \quad
Sumin In\textsuperscript{1} \quad
Suyeon Myeong\textsuperscript{1} \quad
Hogun Park\textsuperscript{2}} \\
\textbf{Sangmin Kim\textsuperscript{3} \quad
Moonhyun Lee\textsuperscript{3} \quad
Daekyeong Park\textsuperscript{3} \quad
Sangpil Kim\textsuperscript{1}}\thanks{Corresponding author.} \\
\textsuperscript{1}Korea University, Republic of Korea \\
\textsuperscript{2}Sungkyunkwan University, Republic of Korea \\
\textsuperscript{3}Hanwha Systems, Republic of Korea \\
\textsuperscript{1}\texttt{\{broiron,ism0705,tnduss,spk7\}@korea.ac.kr} \\
\textsuperscript{2}\texttt{hogunpark@skku.edu} \\
\textsuperscript{3}\texttt{\{smkim0153,moonhyun.lee,daekyeong.park\}@hanwha.com}
}

\begin{document}
\maketitle

\begin{abstract}
Defense is a knowledge-intensive domain that requires precise understanding of specialized terminology, doctrinal concepts, operational procedures, and evolving military events.
Although recent work has explored language technologies for military applications, existing efforts remain fragmented: they are often task-specific, rely on limited adaptation pipelines, or lack comprehensive defense-domain evaluation. 
In this paper, we present \textbf{SALUTE}, an end-to-end framework for benchmarking and adapting LLMs for the defense domain. 
SALUTE integrates Salute-Corpus, a curated corpus from open-access U.S. military doctrine and government documents; Salute-Conv, a grounded instruction dataset from doctrinal sources and decade-long defense news; Salute-Pref, a defense-aware preference dataset; and Salute-Bench, a rigorously filtered benchmark for evaluating defense-domain understanding and reasoning over doctrine and defense news.
Based on these resources, we train Salute-LLM through multi-stage post-training with continual pretraining, supervised fine-tuning, and preference alignment.
Extensive experiments show that Salute-LLM achieves strong defense-domain performance while retaining competitive general capabilities, demonstrating the effectiveness of SALUTE as an end-to-end framework for defense-domain LLM adaptation.
\end{abstract}
\section{Introduction}
\label{sec:intro}

Modern defense environments are increasingly shaped by information overload~\cite{meerveld2023irresponsibility}. 
Defense-relevant information spans doctrine, technical manuals, and rapidly evolving operational news, each requiring precise understanding of specialized terminology, operational context, and domain-specific reasoning.
This motivates the development of language models capable of defense-specific understanding and reasoning.
However, despite their strong general capabilities, general-purpose LLMs~\cite{grattafiori2024llama, yang2025qwen3} remain limited in specialized domains that require precise terminology, domain-specific background knowledge, and reliable reasoning~\cite{shi2025continual}.
These challenges highlight the need for defense-specialized LLMs and an integrated framework for data construction, model adaptation, and comprehensive evaluation.

\begin{figure}[t]
  \includegraphics[width=\columnwidth]{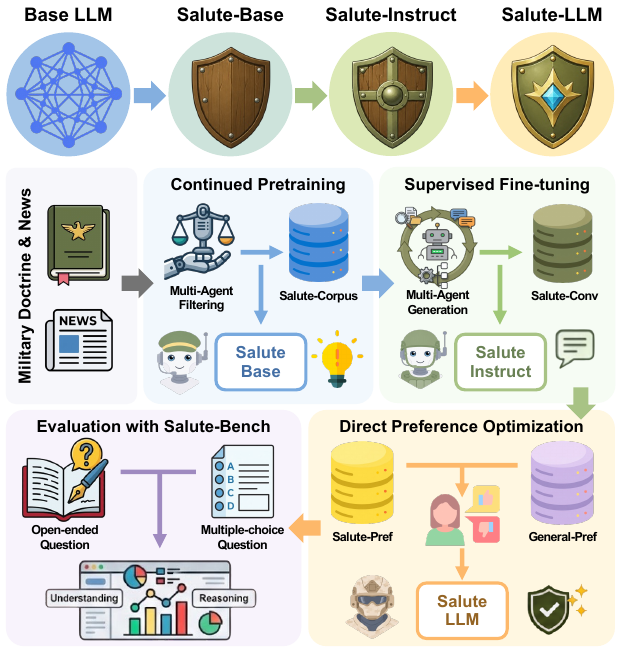}
  \caption{Overview of \textbf{SALUTE}, a unified framework for defense-domain LLM adaptation.}
  \label{fig:teaser_1}
\vspace{-2.0em}
\end{figure}

Recent work has explored language technologies for defense and military applications, but existing efforts still exhibit three important limitations.
\textbf{First}, prior resources~\cite{zhu2024cmnee, palnitkar2026mil} often target specific task settings, such as military event extraction or geospatial planning, rather than broad defense-domain language understanding.
\textbf{Second}, defense-domain LLM efforts~\cite{li2022mlrip, xue2024milchat, ruiz2024fine, fitzgerald2025edgerunner} typically rely on single-stage adaptation and do not provide reproducible pipelines that connect data construction, multi-stage training, and evaluation.
\textbf{Third}, existing defense-domain benchmarks~\cite{hallapy2023milglue, GovBench:JointStaffBench, li2026warbench} remain limited in scope, openness, or coverage, making it difficult to evaluate defense capabilities across stable doctrine and evolving defense events.

To address these limitations, we propose \textbf{SALUTE}, an end-to-end framework for benchmarking and adapting LLMs for the defense domain. 
As shown in Figure~\ref{fig:teaser_1}, SALUTE integrates defense-domain resource construction, multi-stage post-training, and diversified evaluation.

Specifically, we first construct \textbf{Salute-Corpus} from open-access U.S. military doctrine and related government documents to provide the knowledge foundation for continual pretraining. 
To make heterogeneous PDF-based sources suitable for training, we convert them into structured Markdown, segment them into section-aware chunks, remove near-duplicates, and filter low-quality content using multi-agent scoring and a ModernBERT-based~\cite{warner2025smarter} quality model.

We then build \textbf{Salute-Conv}, a defense-domain instruction dataset from doctrinal sources and a decade of defense news, to enable military question answering and domain-specific instruction following. 
Instruction data are generated through task planning, chunk-grounded question generation, retrieval-augmented evidence selection, and grounded answer synthesis, covering both stable doctrinal knowledge and dynamic event-driven defense information. 
We further construct \textbf{Salute-Pref}, a defense-domain preference replay dataset, to preserve domain-specific behaviors during preference alignment.
Based on these resources, \textbf{Salute-LLM} is trained through multi-stage post-training, where continual pretraining injects defense knowledge, supervised fine-tuning adapts it for instruction following, and preference alignment improves response quality. 
Finally, we introduce \textbf{Salute-Bench}, a rigorously filtered benchmark for evaluating defense-domain understanding and reasoning over doctrine and defense news in open-ended and multiple-choice formats.

Extensive experiments show that \textbf{Salute-LLM} improves defense-domain performance while retaining competitive general capabilities.
Our contributions are summarized as follows:

\begin{itemize}
    \item We present \textbf{SALUTE}, a unified framework for defense-domain LLM adaptation across data construction, multi-stage training, and evaluation.

    \item We construct \textbf{Salute-Corpus}, \textbf{Salute-Conv}, and \textbf{Salute-Pref} to support defense-domain pretraining, instruction tuning, and preference alignment.

    \item We introduce \textbf{Salute-Bench}, a rigorously filtered benchmark for evaluating defense-domain understanding and reasoning.
    
    \item Through extensive experiments, we demonstrate that \textbf{Salute-LLM} achieves strong defense-domain performance while retaining competitive general capabilities.

\end{itemize}
\section{Related Work}
\label{sec:related_work}

\begin{table*}[t]
\centering
\scriptsize
\setlength{\tabcolsep}{4.0pt}
\renewcommand{\arraystretch}{0.95}
\resizebox{0.96\textwidth}{!}{%
\begin{tabular}{
@{}
>{\raggedright\arraybackslash}p{2.45cm}
>{\raggedright\arraybackslash}p{3.05cm}
>{\centering\arraybackslash}p{1.25cm}
>{\centering\arraybackslash}p{1.10cm}
>{\centering\arraybackslash}p{0.75cm}
>{\centering\arraybackslash}p{0.75cm}
>{\centering\arraybackslash}p{0.75cm}
>{\centering\arraybackslash}p{1.45cm}
>{\centering\arraybackslash}p{1.25cm}
@{}
}
\toprule
\multirow{2}{*}{\textbf{Category}}
& \multirow{2}{*}{\textbf{Work}}
& \multirow{2}{*}{\makecell[c]{\textbf{Domain}\\[-0.35ex]\textbf{Train Data}}}
& \multirow{2}{*}{\makecell[c]{\textbf{Instruction}\\[-0.35ex]\textbf{Data}}}
& \multicolumn{3}{c}{\textbf{Post-training}}
& \multirow{2}{*}{\makecell[c]{\textbf{Multi-aspect}\\[-0.35ex]\textbf{Evaluation}}}
& \multirow{2}{*}{\makecell[c]{\textbf{Multi}\\[-0.35ex]\textbf{Source}}} \\[-0.4ex]
\cmidrule(lr){5-7}
& & & & \textbf{CPT} & \textbf{SFT} & \textbf{DPO} & & \\[-0.3ex]
\midrule

\textit{Info. Structuring}
& CMNEE~\citeyearpar{zhu2024cmnee}
& \checkmark & -- & -- & -- & -- & -- & -- \\
& MLRIP~\citeyearpar{li2022mlrip}
& \checkmark & -- & \checkmark & -- & -- & -- & \checkmark \\
\midrule

\textit{LLM Domain Adaptation}
& MilChat~\citeyearpar{xue2024milchat}
& \checkmark & \checkmark & -- & \checkmark & -- & -- & -- \\
& TRACLM~\citeyearpar{ruiz2024fine}
& \checkmark & \checkmark & \checkmark & \checkmark & -- & -- & -- \\
& EdgeRunner 20B~\citeyearpar{fitzgerald2025edgerunner}
& \checkmark & \checkmark & -- & \checkmark & -- & -- & \checkmark \\
\midrule

\textit{Benchmark}
& MilGLUE~\citeyearpar{hallapy2023milglue}
& \checkmark & -- & \checkmark & -- & -- & -- & \checkmark \\
& JointStaffBench~\citeyearpar{GovBench:JointStaffBench}
& -- & -- & -- & -- & -- & \checkmark & -- \\
& WARBENCH~\citeyearpar{li2026warbench}
& -- & -- & -- & -- & -- & -- & \checkmark \\
\midrule

\textit{Framework}
& \textbf{SALUTE}
& \checkmark & \checkmark & \checkmark & \checkmark & \checkmark & \checkmark & \checkmark \\
\bottomrule
\end{tabular}
}
\vspace{-.5em}
\caption{
Comparison with prior defense-domain NLP studies.
Columns indicate coverage of domain training data, instruction data, multi-stage post-training, multi-aspect evaluation, and multi-source data construction.
$\checkmark$ and ``--'' denote full and no coverage, respectively.
}
\vspace{-1.5em}
\label{tab:related_comparison}
\end{table*}
\subsection{Domain Adaptation of LLMs}
\label{sec:dapt_llms}
Adapting pretrained language models to specialized domains is an effective way to improve performance under domain-specific terminology, knowledge, and data distributions~\cite{gururangan-etal-2020-dont}. 
Recent efforts have developed domain-specialized LLMs for medicine~\cite{acikgoz2024hippocrates,xie2024me, yang2024pediatricsgpt}, science~\cite{li2025scilitllm,bi2024oceangpt, prabhakar2025omniscience}, law~\cite{colombo2024saullmb, niklaus2025lawinstruct}, finance~\cite{ke2025demystifying, ying-etal-2025-data}, and cybersecurity~\cite{suryanto2026redsage}.
Beyond continual pretraining, these works highlight the importance of domain-specific corpora, instruction data, multi-stage training, and dedicated benchmarks for reliable domain adaptation. 
However, such integrated adaptation and evaluation for the defense-domain remain limited.
Our work addresses this gap by constructing, adapting, and evaluating LLMs for defense-domain tasks.

\subsection{NLP in the Defense Domain}
\label{sec:defense_nlp}
NLP research in the defense domain has mainly explored information structuring, LLM adaptation, and domain-specific evaluation. 
For information structuring, CMNEE~\cite{zhu2024cmnee} constructs a document-level event extraction dataset from military news, while MLRIP~\cite{li2022mlrip} develops a military language representation model with factual and professional knowledge. 
These studies highlight the need for military-specific resources and representations, but focus on structured prediction or representation learning rather than end-to-end LLM adaptation.
Recent work has adapted LLMs to defense settings using military equipment data~\cite{xue2024milchat}, doctrine~\cite{ruiz2024fine}, and task-specific defense data~\cite{fitzgerald2025edgerunner}. 
However, these efforts are often limited to a specific data source, task setting, or adaptation stage, leaving end-to-end defense-domain adaptation underexplored.
Defense-domain benchmarks such as MilGLUE~\cite{hallapy2023milglue}, JointStaffBench~\cite{GovBench:JointStaffBench}, and WARBENCH~\cite{li2026warbench} assess military knowledge and tactical decision-making.
However, they mainly assess static knowledge or scenario-based reasoning, with limited coverage of evolving defense events, leaving multi-source and multi-aspect evaluation underexplored.
SALUTE addresses these gaps through a unified framework that integrates doctrine, government documents, and defense news for data construction, multi-stage adaptation, and evaluation, as shown in Table~\ref{tab:related_comparison}.

\section{SALUTE}
\label{sec:method}

\begin{figure*}[t]
  \includegraphics[width=\linewidth]{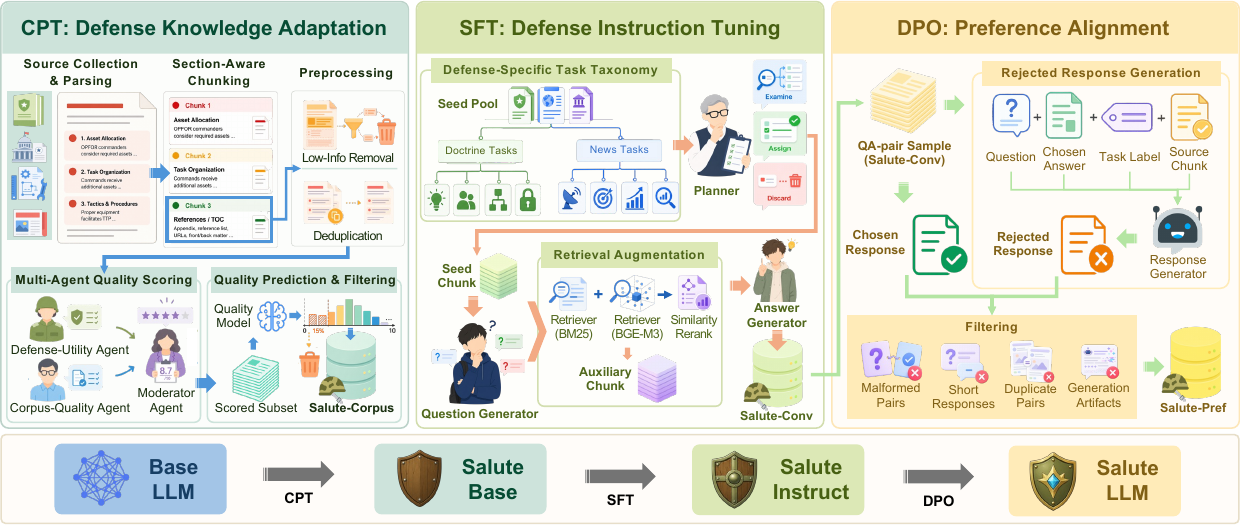}
  \caption{
     End-to-End SALUTE Framework.
    SALUTE builds defense-domain resources and adapts LLMs through CPT, SFT, and DPO, producing \textbf{Salute-Base}, \textbf{Salute-Instruct}, and \textbf{Salute-LLM}.
    }
  \label{fig:main_pipeline}
\vspace{-1.5em}
\end{figure*}

In this section, we describe \textbf{SALUTE}, an end-to-end framework for defense-domain LLM adaptation. 
We first present the construction of three training resources: \textbf{Salute-Corpus} for continual pretraining in Section~\ref{sec:salute_corpus}, \textbf{Salute-Conv} for supervised fine-tuning in Section~\ref{sec:salute_conv}, and \textbf{Salute-Pref} for preference alignment in Section~\ref{sec:salute_preference}. 
We then describe the multi-stage training of \textbf{Salute-LLM} in Section~\ref{sec:salute_llm}, followed by \textbf{Salute-Bench} for evaluating defense-domain knowledge and reasoning in Section~\ref{sec:salute_bench}. 
Figure~\ref{fig:main_pipeline} illustrates the main data construction and training pipeline.

\subsection{Salute-Corpus}
\label{sec:salute_corpus}
A high-quality domain corpus is essential for adapting LLMs to specialized domains, as continual pretraining relies on sufficient coverage of domain terminology, concepts, and discourse patterns~\cite{gururangan-etal-2020-dont}. 
To inject defense-domain knowledge into LLMs, we construct a large-scale defense corpus from open-access military and government documents and refine it through deduplication and fine-grained quality filtering.

\paragraph{Source Collection.}
Defense-domain knowledge covers military concepts, operational procedures, organizational roles, technical guidance, and administrative information. 
Since curated, textbook-quality data improves language model training~\cite{gunasekar2023textbooks}, we use doctrinal and government publications as authoritative, structured defense knowledge sources.
We compile publicly available U.S. defense publications from the Army Publishing Directorate, Marine Corps Publications Electronic Library, U.S. Air Force Doctrine, U.S. Space Force Doctrine, and Department of War publications.
The resulting raw collection contains 8,119 documents and 126.46M tokens.
Additional details on source collection and source-level statistics are provided in Appendix~\ref{sec:supp_source_collect}.

\paragraph{Preprocessing.} 
After source collection, we convert the collected PDF documents into Markdown using Marker~\cite{paruchuri2025marker}. This conversion preserves document structure such as headings and section boundaries, which allows us to perform section-aware chunking rather than splitting documents purely by length. We then remove low-information sections that are unlikely to contain substantive defense knowledge, such as tables of contents, reference lists, and other front/back matter. 
Finally, we apply MinHash-LSH near-duplicate detection to remove redundant chunks across documents. This preprocessing pipeline yields a deduplicated corpus of approximately 121M tokens.

\paragraph{Multi-Agent Quality Filtering.}
Filtering large-scale pretraining corpora requires scalable quality assessment. 
Following recent model-based filtering pipelines~\cite{penedo2024fineweb, henriksson-etal-2025-finerweb}, we score a subset of chunks and train a lightweight quality model for corpus-wide filtering.
Instead of relying on a single LLM judge, we adopt a multi-agent scoring pipeline with complementary perspectives.
A \textit{Defense-Utility Agent} assesses the usefulness of a chunk for defense-domain knowledge acquisition, while a \textit{Corpus-Quality Agent} evaluates general pretraining quality, including coherence, informativeness, readability, and noise. 
A \textit{Moderator Agent} then aggregates the scores and rationales from both agents into a final quality score on a 0-10 scale.
We annotate 15K randomly sampled chunks, train a ModernBERT-based quality model~\cite{warner2025smarter} using these scores as supervision, and apply it to the full deduplicated corpus.
Finally, we exclude the bottom 15\% of chunks according to the predicted quality scores, yielding the final \textbf{Salute-Corpus} with approximately 107M tokens. 
More details on multi-agent scoring, prompt templates, and ModernBERT quality modeling are provided in Appendix~\ref{sec:supp_multiagent_filtering}.
Figure~\ref{fig:corpus_graph} shows the distribution of predicted quality scores.


\subsection{Salute-Conv}
\label{sec:salute_conv}

While Salute-Corpus provides the knowledge foundation for continual pretraining, defense-domain LLMs also require supervised examples for military question answering and domain-specific instruction following. 
To bridge this gap, we construct a defense-domain conversation dataset for supervised fine-tuning by converting defense text into diverse question-answer and instruction-following examples across multiple defense task categories.

\paragraph{Seed construction.} 
Doctrinal documents provide stable and authoritative defense knowledge, but they do not fully capture emerging military events and dynamic developments, such as deployments, exercises, capability updates, and international cooperation.
To complement doctrinal knowledge with timely defense information, we collect publicly accessible defense news and press releases from official military and government sources over a 10-year period, totaling 36.99M tokens.
To balance doctrinal and news-based sources in instruction generation, we randomly sample 10\% of the doctrinal chunks and combine them with the collected news articles to form the seed pool.

\paragraph{Task-Guided Instruction Generation.}
After seed construction, we define a defense-specific task taxonomy to guide instruction-data generation, as summarized in Table~\ref{tab:task_taxonomy}.
The taxonomy is designed to cover both stable doctrinal knowledge and dynamic event-driven defense knowledge through doctrine-oriented and news-oriented tasks.
For each seed chunk, a \textit{planner} examines the chunk content and either assigns the most appropriate task type or discards the chunk if it does not support any predefined task.
Conditioned on the selected task and seed chunk, a \textit{question generator} then produces multiple questions grounded in the chunk content.
Further details on task planning and question generation are provided in Appendix~\ref{sec:supp_conv_planning}.

\begin{figure}[t]
    \centering
    \includegraphics[width=\linewidth]{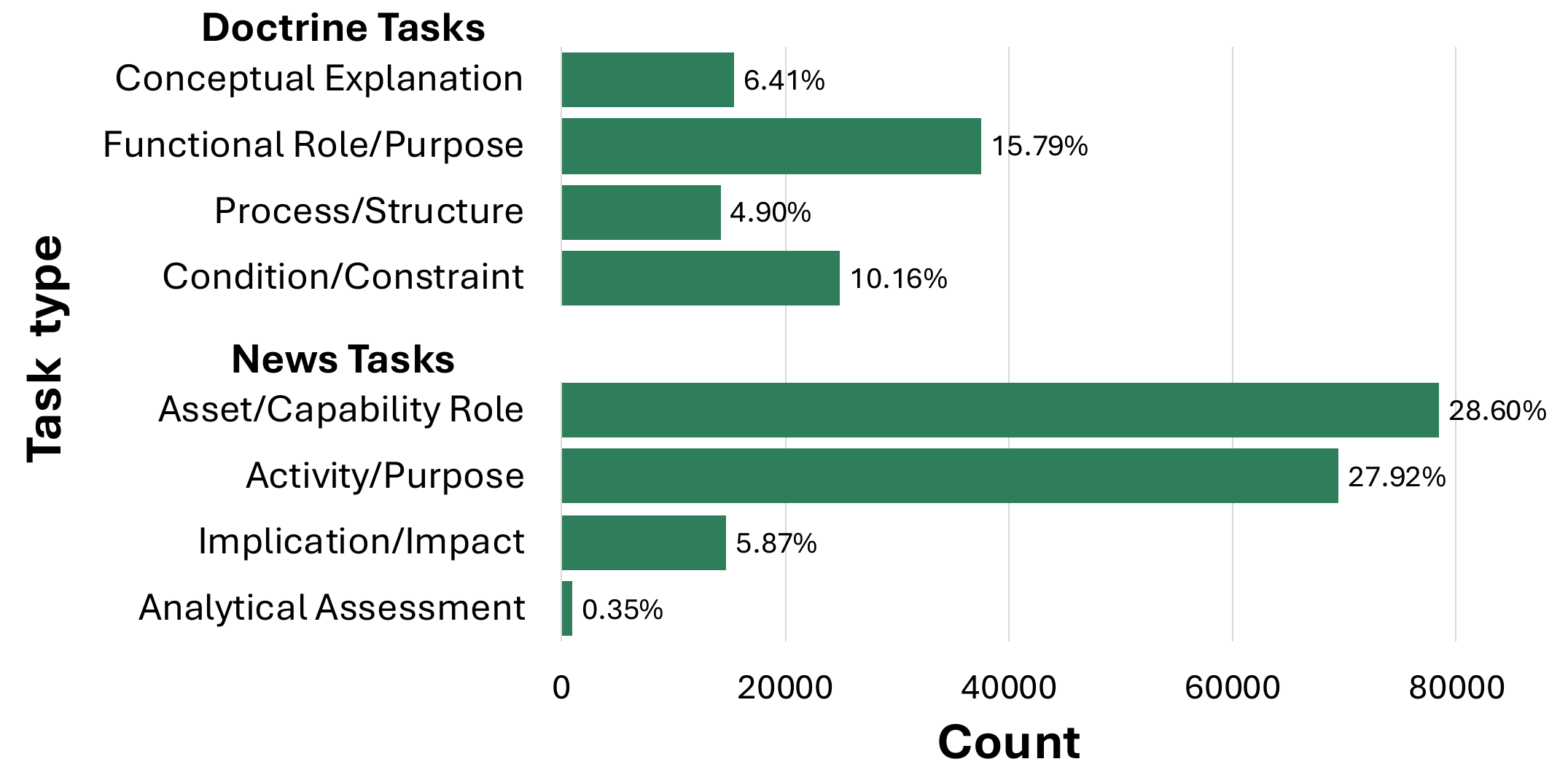}
    \caption{
     Task distribution of \textbf{Salute-Conv}.
     Bars show the number and proportion of question-answer pairs for each doctrine and defense news task.
     }
    \label{fig:graph2}
    \vspace{-1.5em}
\end{figure}

\paragraph{Retrieval-Augmented Answer Synthesis.}
To improve answer factuality, we synthesize responses with retrieval augmentation, inspired by recent hybrid retrieval approaches~\cite{wang-etal-2024-searching,yu2024rankrag}.
For each generated question, the original seed chunk is treated as primary evidence, while auxiliary chunks are retrieved only to support, clarify, or refine the answer. 
Auxiliary evidence is obtained through hybrid retrieval with BM25~\cite{robertson2009probabilistic} and BGE-M3~\cite{chen-etal-2024-m3}, followed by reranking based on similarity to the seed chunk using Qwen3-Embedding~\cite{zhang2025}.
The \textit{answer generator} is then prompted to prioritize the seed chunk over auxiliary evidence.
Further details on retrieval and answer generation are provided in Appendix~\ref{sec:supp_conv_rag}.
This evidence-grounded pipeline mitigates hallucinated or weakly supported responses and yields \textbf{Salute-Conv}, a 255K-pair defense-domain instruction dataset.
Figure~\ref{fig:graph2} shows the task distribution, and Appendix~\ref{sec:supp_conv_statistics} provides additional statistics and examples.

\begin{figure}[t]
    \centering
    \includegraphics[width=\linewidth]{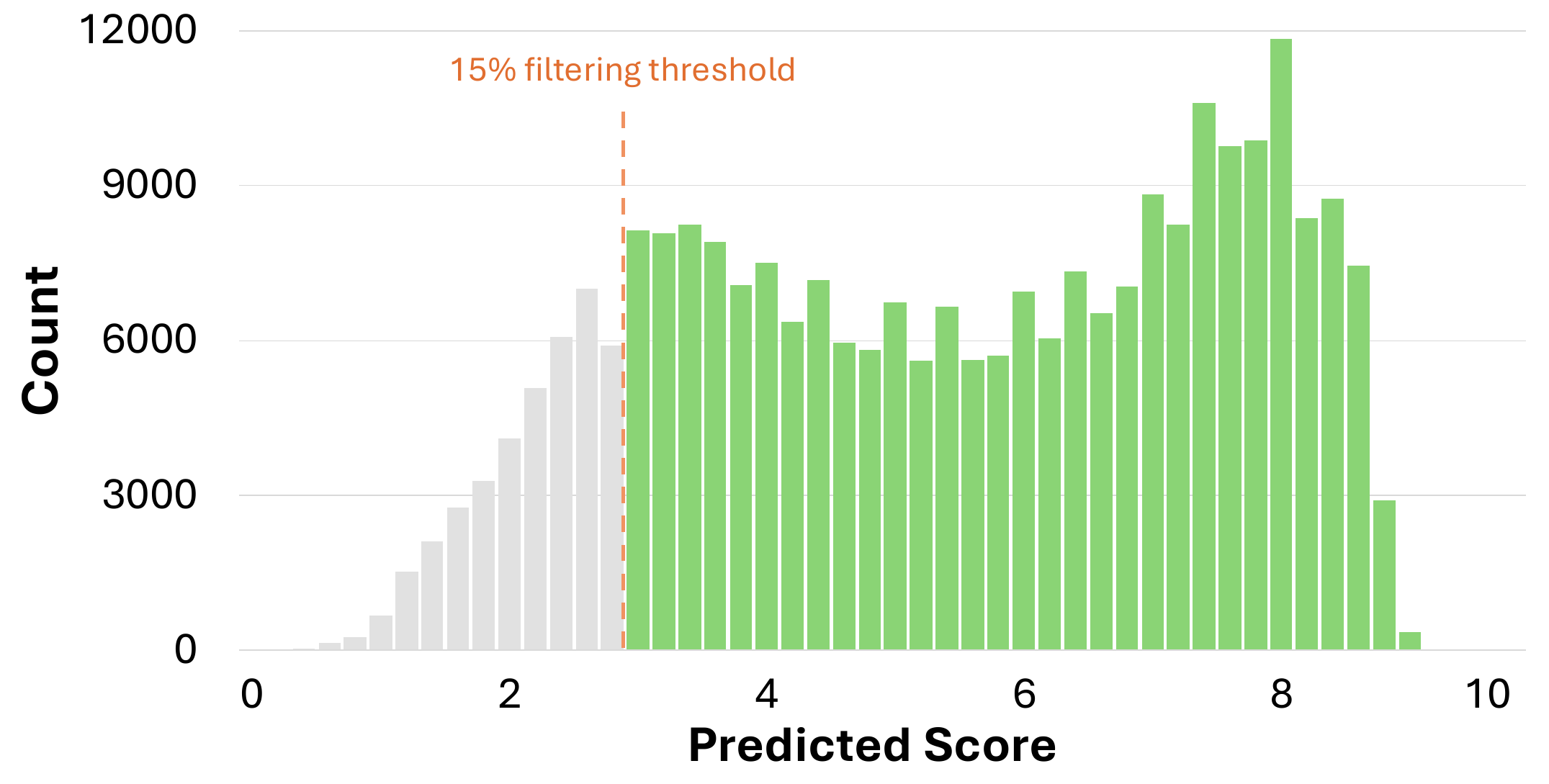}
    \caption{
    Predicted quality score distribution.
    Gray bars denote the filtered bottom 15\%; green bars denote chunks retained for \textbf{Salute-Corpus}.
    }
    \label{fig:corpus_graph}
    \vspace{-1.5em}
\end{figure}

\subsection{Salute-Pref}
\label{sec:salute_preference}
Preference alignment improves response quality, but general-domain preference data alone may dilute domain-specific behaviors. 
To preserve defense-domain response patterns, we construct \textbf{Salute-Pref}, a defense-aware replay dataset for the final preference alignment stage.
Since the quality of chosen responses is important for effective preference optimization~\cite{pan2026what}, we sample 20K question-answer pairs from the \textbf{Salute-Conv} training split and use their source-grounded answers as chosen responses.
For each instance, we generate a fluent but lower-quality rejected response with Qwen3-30B~\cite{yang2025qwen3}, conditioned on the question, source chunk, task label, and chosen answer. 
Rejected responses contain realistic defects such as incompleteness, over-generalization, weak grounding, or subtle domain-specific confusion.
After removing malformed pairs, overly short responses, near-duplicates, and meta-generation artifacts, we obtain approximately 19K preference pairs. 
Appendix~\ref{sec:supp_pref} provides further details on rejected-response generation, filtering criteria, and qualitative examples.

\subsection{Salute-LLM}
\label{sec:salute_llm}
We train \textbf{Salute-LLM} using LlamaFactory~\cite{zheng-etal-2024-llamafactory} with full-parameter optimization. 
Our training pipeline consists of three stages: continual pretraining for domain knowledge acquisition, supervised fine-tuning for instruction following, and preference optimization for response alignment.

Starting from Qwen3-8B-Base~\cite{yang2025qwen3}, we perform continual pretraining on a mixture of Salute-Corpus and 10.2M FineWeb~\cite{penedo2024fineweb} replay tokens. 
All Salute-Bench source chunks are excluded from the Salute-Corpus training split.
We train for one epoch with a learning rate of $8\times10^{-6}$, cosine scheduling, and a warmup ratio of 0.03, yielding \textbf{Salute-Base}.

We then conduct supervised fine-tuning to convert the acquired domain knowledge into instruction-following behavior. 
Since the composition of instruction data affects the balance among model capabilities~\cite{dong-etal-2024-abilities}, we adopt a two-stage schedule that shifts from general instruction-following preservation to defense-domain specialization. 
Stage 1 mixes 255K Salute-Conv with 939K examples from the Tulu-3-SFT-Mixture~\cite{lambert2025tulu}, while Stage 2 keeps the same Salute-Conv data but reduces the replay set to 100K sampled Tulu examples. 
The two stages use learning rates of $3\times10^{-6}$ and $1.0\times10^{-6}$, respectively, with linear scheduling and a warmup ratio of 0.03, and are trained for two and one epochs, yielding \textbf{Salute-Instruct}.

\begin{table}[t]
\centering
\small
\setlength{\tabcolsep}{3pt}
\begin{tabular}{lp{4.2cm}}
\toprule
\textbf{Task} & \textbf{Capability} \\
\midrule
\multicolumn{2}{l}{\textit{Doctrine Tasks}} \\
Conceptual Explanation & Doctrinal concepts, principles, and distinctions. \\
Functional Role/Purpose & Functions and purposes of doctrines, roles, and units. \\
Process/Structure & Doctrinal steps, structures, and processes. \\
Condition/Constraint & Conditions, constraints, exceptions, and decision criteria. \\
\midrule
\multicolumn{2}{l}{\textit{News Tasks}} \\
Asset/Capability Role & Roles and significance of assets, systems, and capabilities. \\
Activity/Purpose & Purposes of deployments, exercises, launches, and cooperation. \\
Implication/Impact & Military implications, strategic signals, and operational impacts. \\
Analytical Assessment & Factors, indicators, uncertainties, and assessment criteria. \\
\bottomrule
\end{tabular}
\caption{
Defense task taxonomy guiding task planning and question generation in Salute-Conv construction.
}
\vspace{-1.5em}
\label{tab:task_taxonomy}
\end{table}

Finally, we perform preference alignment with DPO~\cite{rafailov2023direct}. 
We mix 19K Salute-Pref pairs with 273K examples from the Tulu-3-8B Preference Mixture~\cite{lambert2025tulu}, using Salute-Pref as defense-domain replay data to preserve domain-specific behaviors during general preference alignment and mitigate drift from previous stages.
We use a sigmoid preference loss with $\beta=0.1$, training for a single epoch with a learning rate of $5\times10^{-7}$ and a warmup ratio of 0.1.
The resulting model is denoted as \textbf{Salute-LLM}.
Additional details on training resources and implementation details are provided in Appendix~\ref{sec:supp_train}.

\subsection{Salute-Bench}
\label{sec:salute_bench}
To evaluate defense-domain capabilities, we construct \textbf{Salute-Bench} in both open-ended and multiple-choice formats from held-out source chunks reserved during the Salute-Conv construction process. 
We sample 150 chunks from each of the eight task categories in Table~\ref{tab:task_taxonomy}, yielding 1,200 held-out source chunks that are excluded from all model-training data. 
The instruction-generation pipeline produces 8,163 open-ended (OE) QA instances from these chunks, which serve as the initial benchmark pool.
Further details on Salute-Bench are provided in Appendix~\ref{sec:supp_salute_bench}.

\paragraph{Multiple-Choice Question Generation.}

We further convert each open-ended QA instance in the initial benchmark pool into a four-choice multiple-choice question (MCQ) using Qwen3-30B~\cite{yang2025qwen3}.
Given the source chunk, open-ended question, and reference answer, the model transforms the QA pair into a four-choice multiple-choice item with one correct option and three plausible distractors.
Conditioning on the source chunk helps generate context-aware distractors that are close to the correct answer but unsupported or inconsistent with the evidence.
We shuffle the options and update the answer key to reduce answer-position bias, enabling answer-selection evaluation alongside free-form generation.

\begin{table}[t]
\centering
\small
\setlength{\tabcolsep}{3.5pt}
\begin{tabular}{lrrr}
\toprule
\textbf{Task} & \textbf{\#Init.} & \textbf{\#OE} & \textbf{\#MCQ} \\
\midrule
\multicolumn{4}{l}{\textit{Doctrine Tasks}} \\
Conceptual Explanation & 913 & 312 & 114 \\
Functional Role/Purpose  & 951 & 385 & 149 \\
Process/Structure & 1,172 & 419 & 215 \\
Condition/Constraint & 983 & 363 & 170 \\
\midrule
\multicolumn{4}{l}{\textit{News Tasks}} \\
Asset/Capability Role & 1,074 & 344 & 70 \\
Activity/Purpose  & 1,001 & 311 & 65 \\
Implication/Impact & 956 & 163 & 121 \\
Analytical Assessment & 1,113 & 341 & 69 \\
\midrule
\textbf{Total} & \textbf{8,163} & \textbf{2,638} & \textbf{973} \\
\bottomrule
\end{tabular}
\caption{
Task-wise statistics of Salute-Bench after quality filtering.
Init. denotes the initial open-ended QA pool before filtering.
}
\vspace{-1.5em}
\label{tab:salute_bench_stats}
\end{table}

\paragraph{Quality Filtering.}
We apply strict GPT-5-based filtering~\cite{singh2025openai} to retain valid, answerable, source-grounded, and discriminative instances, with all scores assigned on a 1-5 scale.

For open-ended questions, GPT-5 assigns a validity score based on clarity, standalone answerability from the source chunk, and direct support for the reference answer.
Open-ended instances with validity scores below 4 are discarded.

For multiple-choice questions, GPT-5 assigns separate validity and difficulty scores.
Validity assesses source grounding, answer-key correctness, and single-answer consistency.
Difficulty assesses whether the item requires more than simple recall, includes plausible distractors, and avoids shortcut cues such as option-length imbalance, copied wording, or mismatched option types.
MCQ instances are retained only if they score at least 4 on both dimensions.
After filtering, the final Salute-Bench contains 2,638 open-ended questions and 973 multiple-choice questions, with task-wise statistics reported in Table~\ref{tab:salute_bench_stats}.



\begin{table*}[t]
\centering
\small
\setlength{\tabcolsep}{5pt}
\begin{tabular}{lccccc c}
\toprule
\multirow{2}{*}{\textbf{Model}} 
& \multicolumn{5}{c}{\textbf{Open-Ended QA}} 
& \multicolumn{1}{c}{\textbf{Multiple-Choice QA}} \\
\cmidrule(lr){2-6} \cmidrule(lr){7-7}
& \textbf{Correct.} 
& \textbf{Complete.} 
& \textbf{Relevant.} 
& \textbf{Overall} 
& \textbf{Avg.} 
& \textbf{Acc.} \\
\midrule
\multicolumn{7}{l}{\textit{Open-weight baselines}} \\
Qwen3-8B-Base      & 55.35 & 47.73 & 70.27 & 50.74 & 56.02 & 72.04 \\
Qwen3-8B           & 54.70 & 46.95 & 69.34 & 49.85 & 55.21 & 77.28 \\
Llama-3.1-8B-Instruct & 47.29 & 45.77 & 59.11 & 45.72 & 49.47 & 70.20 \\
Ministral-3-8B-Instruct & 53.14 & 48.24 & 63.64 & 48.94 & 53.49 & 77.80 \\
Tulu-3-8B  & 58.59 & 56.31 & 71.90 & 56.52 & 60.83 & 63.00 \\
Granite-3.3-8B-Instruct & 54.48 & 50.85 & 68.67 & 52.03 & 56.51 & 71.63 \\
\midrule
\multicolumn{7}{l}{\textit{SALUTE variants}} \\
Salute-8B-Base     & 58.89 & 54.60 & 73.20 & 56.02 & 60.68 & 78.52 \\
Salute-8B-Instruct & \underline{60.62} & \underline{56.71} & \textbf{78.47} & \underline{58.17} & \underline{63.49} & \underline{85.09} \\
\textbf{Salute-LLM} & \textbf{60.91} & \textbf{59.87} & \underline{77.71} & \textbf{59.53} & \textbf{64.51} & \textbf{86.02} \\
\midrule
\multicolumn{7}{l}{\textit{Strong reference models}} \\
Qwen3-30B & 61.82 & 56.30 & 75.91 & 57.30 & 62.83 & 78.11 \\
GPT-5 & 72.85 & 65.20 & 84.23 & 66.88 & 72.29 & 92.18 \\
\bottomrule
\end{tabular}
\caption{
Main results on Salute-Bench. 
Avg. denotes the average of the four open-ended judge scores, and Acc. denotes multiple-choice QA accuracy.
\textbf{Bold} and \underline{underline} indicate the best and second-best results among 8B-scale open-weight models, excluding strong reference models.
}
\vspace{-1.5em}
\label{tab:salute_bench_results}
\end{table*}

\paragraph{Defense Expert Assessment.}
In addition to LLM-based filtering, 5 practitioners from a defense company audit 400 randomly sampled benchmark instances, including 200 open-ended and 200 multiple-choice questions.
Items are checked for domain appropriateness, factual correctness, source support, standalone clarity, and, for MCQs, single-answer validity and distractor plausibility.
Following our protocol, all audited instances meet the acceptance standard of satisfying all or all but one applicable criterion.
Further details on the audit protocol and criteria are provided in Appendix~\ref{sec:supp_expert_assessment}.

\paragraph{Evaluation.}
For evaluation, MCQ instances are scored by accuracy, while open-ended responses are evaluated by GPT-OSS-120B~\cite{agarwal2025gpt} as a model-as-judge along correctness, completeness, relevance, and overall quality, with raw 1-5 judge scores rescaled to 0-100.
\section{Experiments}
\label{sec:exp}
\subsection{Experimental Setup}
\label{sec:exp_setup}
\paragraph{Baselines.}
We compare \textbf{Salute-LLM} with open-weight LLMs of similar scale to assess the effect of defense-domain adaptation under comparable model capacity. 
The baselines include Qwen3-8B-Base, Qwen3-8B~\cite{yang2025qwen3}, Llama-3.1-8B-Instruct~\cite{grattafiori2024llama}, Tulu-3-8B~\cite{lambert2025tulu}, Ministral-3-8B-Instruct~\cite{liu2026ministral}, and Granite-3.3-8B-Instruct~\cite{granite2024granite}. 
We additionally report results from stronger reference models, including Qwen3-30B and GPT-5~\cite{singh2025openai}, to contextualize the difficulty of Salute-Bench. 
To analyze how each training stage contributes to the final model, we also evaluate intermediate variants, \textbf{Salute-Base} and \textbf{Salute-Instruct}.

\paragraph{Evaluation.} 
For defense-domain evaluation, we use \textbf{Salute-Bench} as the primary benchmark, since existing defense-domain benchmarks often have limited accessibility or narrow scope, as detailed in Appendix~\ref{sec:external_benchmark_availability}.
To assess general capability retention after domain adaptation, we evaluate models with the EleutherAI lm-evaluation-harness~\cite{eval-harness} on MMLU~\cite{hendrycks2021measuring}, TruthfulQA~\cite{lin2022truthfulqa}, ARC~\cite{clark2018think}, IFEval~\cite{zhou2023instruction}, GSM8K~\cite{cobbe2021gsm8k}, and GPQA~\cite{rein2024gpqa}.
IFEval is excluded for base models, as it targets instruction-following ability.
We report individual benchmark scores and macro-averages over applicable benchmarks, with benchmark descriptions and evaluation metrics provided in Appendix~\ref{sec:supp_general_bench}.

\begin{table*}[t]
\centering
\small
\setlength{\tabcolsep}{6pt}
\begin{tabular}{lccccccc}
\toprule
\multirow{2}{*}{\textbf{Model}}
& \multicolumn{7}{c}{\textbf{General LLM Benchmarks}} \\
\cmidrule(lr){2-8}
& \textbf{Mean}
& \textbf{MMLU}
& \textbf{TruthfulQA}
& \textbf{ARC}
& \textbf{IFEval}
& \textbf{GSM8K}
& \textbf{GPQA} \\
\midrule
\multicolumn{8}{l}{\textit{Open-weight baselines}} \\
Qwen3-8B-Base      
& \underline{63.95} & \underline{76.94} & 50.99 & \textbf{64.41} & -- & 85.67 & \underline{41.74} \\
Qwen3-8B  
& 63.53 & 72.04 & 53.16 & 57.16 & \textbf{76.34} & 85.44 & 37.05 \\
Llama-3.1-8B-Instruct 
& 62.72 & 68.66 & 55.04 & 60.49 & \underline{73.75} & 83.16 & 35.26 \\
Ministral-3-8B-Instruct 
& 62.16 & 76.44 & \underline{63.88} & 63.05 & 52.12 & 79.75 & 37.72 \\
Tulu-3-8B
& 62.56 & 62.22 & 60.34 & 55.20 & \textbf{76.34} & \textbf{88.70} & 32.58 \\
Granite-3.3-8B-Instruct 
& 60.86 & 65.16 & \textbf{66.64} & 60.23 & 64.51 & 76.72 & 31.91 \\
\midrule
\multicolumn{8}{l}{\textit{SALUTE variants}} \\
Salute-8B-Base  
& 62.96 & \textbf{76.96} & 48.61 & \underline{63.82} & -- & 84.83 & 40.62 \\
Salute-8B-Instruct  
& 61.72 & 74.69 & 48.90 & 59.47 & 64.87 & 81.57 & 40.84 \\
\textbf{Salute-LLM} 
& \textbf{64.69} & 75.35 & 52.31 & 61.00 & 70.97 & \underline{86.35} & \textbf{42.19} \\
\midrule
\multicolumn{8}{l}{\textit{Strong reference models}} \\
Qwen3-30B 
& 69.10 & 81.00 & 60.04 & 59.98 & 73.75 & 94.76 & 45.08 \\
GPT-5 
& 82.39 & 85.97 & 81.72 & 95.90 & 84.47 & 94.31 & 52.00 \\
\bottomrule
\end{tabular}
\caption{
General benchmark results for assessing general capability retention.
Mean denotes the macro-average over applicable benchmarks.
\textbf{Bold} and \underline{underline} indicate the best and second-best results among 8B-scale open-weight models, excluding strong reference models.
}
\label{tab:general_benchmark_results}
\vspace{-1.5em}
\end{table*}

\subsection{Performance on Salute-Bench}
\label{sec:exp_salute_bench}
Table~\ref{tab:salute_bench_results} reports the main results on Salute-Bench. 
Among 8B-scale open-weight models, \textbf{Salute-LLM} achieves the best performance, with an open-ended average score of 64.51 and an MCQ accuracy of 86.02. 
This indicates gains of 3.68 points in open-ended average over Tulu-3-8B and 8.22 points in MCQ accuracy over Ministral-3-8B-Instruct, the strongest non-SALUTE baselines in each setting.
Notably, Salute-LLM also outperforms the larger Qwen3-30B reference model in both open-ended average and MCQ accuracy.
This suggests that defense-domain adaptation can improve domain-specific reasoning and answer selection beyond the gains from model scale alone.

The SALUTE variants highlight the importance of multi-stage adaptation. 
Performance improves progressively from Salute-8B-Base to Salute-8B-Instruct and Salute-LLM, showing that continual pretraining, supervised fine-tuning, and preference alignment provide complementary benefits. 
We provide task-wise and qualitative results in Appendix~\ref{sec:supp_taskwise} and Appendix~\ref{sec:supp_qualitative}.

\subsection{General Benchmark Performance}
\label{sec:exp_general_bench}

Table~\ref{tab:general_benchmark_results} shows that \textbf{Salute-LLM} retains competitive general capabilities after defense-domain adaptation. 
Among 8B-scale open-weight models, Salute-LLM achieves the highest mean score of 64.69, outperforming Qwen3-8B and other baselines.
It also achieves the best GPQA score among the compared 8B models, indicating that our pipeline improves defense-domain performance while largely preserving general reasoning ability.

The SALUTE variants further show the role of replay and preference alignment. 
\textbf{Salute-8B-Base} remains close to Qwen3-8B-Base, indicating that general replay during continual pretraining helps mitigate forgetting. 
Although supervised fine-tuning lowers the general benchmark average, the final preference alignment stage improves the mean score from 61.72 to 64.69, with gains across all reported benchmarks, including IFEval, GSM8K, and GPQA. 
These results suggest that preference alignment helps recover general instruction-following and reasoning ability while preserving defense-domain specialization.

\subsection{Ablation Study}
\label{sec:ablation}


Table~\ref{tab:ablation} presents an ablation study of the multi-stage adaptation pipeline. 
All variants are initialized from Qwen3-8B-Base, allowing us to isolate the effects of continual pretraining, supervised fine-tuning, and preference alignment. 
CPT-only improves Salute-Bench performance over the base model, indicating that domain-adaptive pretraining provides a useful defense-knowledge foundation. 
The comparison between SFT-only and CPT $\rightarrow$ SFT shows that pretrained defense knowledge provides additional benefits beyond instruction tuning alone.
\begin{table}[ht]
\centering
\small
\setlength{\tabcolsep}{1.8pt}
\begin{tabular}{lcccccc}
\toprule
\textbf{Model} 
& \textbf{CPT} 
& \textbf{SFT} 
& \textbf{DPO} 
& \textbf{OE Avg.} 
& \textbf{MCQ} 
& \textbf{Gen. Avg.} \\
\midrule
Base
& \xmark & \xmark & \xmark & 56.02 & 72.04 & 63.95 \\
CPT-Only
& \cmark & \xmark & \xmark & 60.68 & 78.52 & 62.96 \\
SFT-Only
& \xmark & \cmark & \xmark & 62.12 & 80.26 & 61.14 \\
CPT $\rightarrow$ SFT
& \cmark & \cmark & \xmark & 63.49 & 85.09 & 61.72 \\
SFT $\rightarrow$ DPO
& \xmark & \cmark & \cmark & 63.54 & 84.58 & 64.58 \\
Full
& \cmark & \cmark & \cmark & \textbf{64.51} & \textbf{86.02} & \textbf{64.69} \\
\bottomrule
\end{tabular}
\caption{
Ablation study of the multi-stage adaptation pipeline.
OE Avg. and MCQ are measured on Salute-Bench, while Gen. Avg. denotes the mean score across general benchmarks.
}
\vspace{-1.5em}
\label{tab:ablation}
\end{table}
Comparing CPT $\rightarrow$ SFT with the full pipeline shows that DPO further improves both Salute-Bench performance and general benchmark average. 
The SFT $\rightarrow$ DPO variant also shows that preference alignment without CPT is helpful, but the full pipeline performs best across all metrics. 
These results show that each stage contributes complementary benefits: CPT builds the domain knowledge foundation, SFT operationalizes it for defense-domain instructions, and DPO improves alignment while preserving defense capabilities.
Additional replay and filtering-threshold ablations are provided in Appendix~\ref{sec:supp_replay_ablation} and Appendix~\ref{sec:supp_filtering_ablation}.


\section{Conclusion}
\label{sec:conclusion}
We presented \textbf{SALUTE}, a unified resource and evaluation framework for defense-domain LLM adaptation. 
SALUTE contributes \textbf{Salute-Corpus}, \textbf{Salute-Conv}, \textbf{Salute-Pref}, and \textbf{Salute-Bench}, supporting the full pipeline from data construction and model adaptation to benchmark-based evaluation.
Experiments show that \textbf{Salute-LLM} achieves strong defense-domain performance while retaining competitive general capabilities. 
More broadly, our findings highlight that reliable defense-domain LLM adaptation requires integrated resources, including curated corpora, grounded instruction data, preference replay, and rigorously filtered benchmarks. 
We hope SALUTE provides a foundation for more systematic and reproducible research on defense-specialized language models.

\section{Limitations}
\label{sec:supp_limit}
While SALUTE provides an end-to-end framework for defense-domain LLM adaptation and evaluation, its scope remains bounded. 
First, because our corpus and benchmark are built from open-access doctrine, government publications, and defense news, SALUTE primarily reflects publicly available, English-language, and largely U.S.-centric defense knowledge. 
Thus, non-U.S. doctrines, multilingual contexts, and classified procedures are less covered.
Second, SALUTE adopts automated generation, filtering, and evaluation pipelines to enable scalable resource construction. 
Although we ground generation in source evidence and apply retrieval augmentation, multi-agent filtering, and quality control, future work may further strengthen the pipeline with broader human expert validation.
Finally, our experiments focus on 8B-scale open-weight models and a specific CPT--SFT--DPO pipeline. 
Future work should examine broader model scales, base model families, multilingual sources, and evaluation settings.


\section{Ethics Statement}
\label{sec:ethics}
SALUTE is built exclusively from open-access defense-domain sources, including public doctrine, government publications, and defense news, as summarized in Table~\ref{tab:source_statistics}. 
We use these sources in accordance with their stated licenses and terms of use, and do not use classified, restricted, or intentionally collected personally sensitive information. 
SALUTE tasks are designed to evaluate defense-domain knowledge, rather than identifying private individuals or inferring sensitive personal attributes.
For all human evaluations in this work, evaluators were recruited as domain practitioners through an external defense-domain organization. 
No crowdsourcing platform was used, and no separate per-item participant payment was provided by the authors.
Evaluators were informed that their judgments would be used in aggregate for research and evaluation purposes, and no personally identifying information about the evaluators is reported.
As SALUTE incorporates defense-domain knowledge, it may carry potential dual-use risks. 
We therefore release SALUTE strictly for research and evaluation purposes, and do not intend it for real-world operations, mission-critical settings, or autonomous decision support. 
Users should respect the original data licenses, avoid attempts to infer sensitive or non-public information, and apply the resources with appropriate human oversight.
To support responsible use, we plan to release the models, datasets, and code through a gated-access process under research-use-only terms with explicit prohibited-use clauses, subject to applicable data licenses and release policies.
We used AI assistants only for language polishing, writing support at sentence level, and limited coding/debugging assistance. 

\section*{Acknowledgement}
This work was supported by a grant-in-aid of HANWHA SYSTEMS (96\%), Sports and Tourism R\&D Program through the Korea Creative Content Agency grant funded by the Ministry of Culture, Sports and Tourism (International Collaborative Research and Global Talent Development for the Development of Copyright Management and Protection Technologies for Generative AI, RS-2024-00345025, 1\%), the National Research Foundation of Korea(NRF) grant funded by the Korea government(MSIT) (RS-2025-00521602, 1\%), Institute of Information \& Communications Technology Planning \& Evaluation (IITP) grant funded by the Korea government (MSIT) (No. RS-2019-II190079, Artificial Intelligence Graduate School Program (Korea University), 1\%), and the Advanced GPU Utilization Support Program funded by the Government of the Republic of Korea (Ministry of Science and ICT).

\bibliography{main}

\clearpage
\appendix

\section*{Appendix}
\label{sec:appendix}
This appendix provides supplementary details and analyses supporting the main paper. 
We first provide additional details on Salute-Corpus, including source collection, corpus filtering, multi-agent scoring, and ModernBERT-based quality modeling (Section~\ref{sec:supp_corpus}). 
We further describe the Salute-Conv generation pipeline, including task-wise examples, retrieval-augmented answer generation, prompts, and dataset statistics (Section~\ref{sec:supp_conv}). 
We present the construction and filtering process of Salute-Pref (Section~\ref{sec:supp_pref}). 
We then report the training setup and computational resources used for Salute-LLM (Section~\ref{sec:supp_train}). 
Finally, we provide evaluation details (Section~\ref{sec:supp_evaluation}) and additional experimental results (Section~\ref{sec:more_exp}), including replay and filtering ablations, task-wise and qualitative analyses, a retrieval-augmented generation baseline, and human validation.

\section{Details of Salute-Corpus}
\label{sec:supp_corpus}

\begin{table*}[t]
\centering
\small
\setlength{\tabcolsep}{5pt}
\begin{tabular}{llrr}
\toprule
\textbf{Source} & \textbf{Provider} & \textbf{\#Docs and Articles} & \textbf{Tokens} \\
\midrule
\multicolumn{4}{l}{\textit{Doctrine and government publications}} \\
Army Publishing Directorate & U.S. Army & 4,346 & 44.78M \\
Marine Corps Publications Electronic Library & U.S. Marine Corps & 1,366 & 52.18M \\
U.S. Air Force Doctrine & U.S. Air Force & 1,791 & 25.11M \\
Department of War Publications & U.S. Department of War & 579 & 4.24M \\
U.S. Space Force Doctrine & U.S. Space Force & 37 & 0.15M \\
\textbf{Subtotal} &  & \textbf{8,119} & \textbf{126.46M} \\
\midrule
\multicolumn{4}{l}{\textit{Defense news and press releases}} \\
Army Worldwide News & U.S. Army & 326 & 0.28M \\
Joint Chiefs of Staff News & Joint Chiefs of Staff & 1,296 & 0.56M \\
U.S. Department of War News & U.S. Department of War & 12,447 & 7.70M \\
GOV.UK Defence and Armed Forces News & UK Ministry of Defence & 9,460 & 8.68M \\
U.S. Space Force News & U.S. Space Force & 1,047 & 0.39M \\
U.S. Air Force News & U.S. Air Force & 43,574 & 4.91M \\
U.S. Marine Corps News & U.S. Marine Corps & 2,550 & 1.44M \\
U.S. Navy News & U.S. Navy & 18,317 & 13.04M \\
\textbf{Subtotal} &  & \textbf{89,017} & \textbf{36.99M} \\
\midrule
\textbf{Total} &  & \textbf{97,136} & \textbf{163.45M} \\
\bottomrule
\end{tabular}
\caption{
Statistics of the SALUTE source collections.
We collect open-access U.S. military doctrine, government publications, together with official defense news sources.
Token counts are computed using the Qwen3-8B tokenizer.
}
\vspace{-1.0em}
\label{tab:source_statistics}
\end{table*}

\subsection{Source Collection}
\label{sec:supp_source_collect}
We collect Salute-Corpus from open-access military doctrine and government publication repositories. 
The source collection focuses on public documents that contain structured defense-domain knowledge, including doctrinal concepts, operational procedures, organizational roles, technical guidance, and administrative instructions. 
In total, we compile 8,119 doctrine and government documents, comprising 126.46M tokens.
Detailed source statistics are reported in Table~\ref{tab:source_statistics}, and public source URLs are provided in Table~\ref{tab:source_urls}.
We prioritize doctrine and government publications because they provide relatively authoritative and terminology-rich text suitable for continual pretraining.

\subsection{Multi-Agent Quality Filtering}
\label{sec:supp_multiagent_filtering}

\paragraph{Multi-agent scoring.}
To obtain scalable supervision for corpus filtering, we randomly sample 15K chunks from the deduplicated corpus and score them with a multi-agent evaluation pipeline. 
The pipeline consists of three agents: \textit{Defense-Utility Agent}, \textit{Corpus-Quality Agent}, and a \textit{Moderator Agent}. 
The \textit{Defense-Utility Agent} evaluates defense-domain utility along three dimensions: defense relevance, doctrinal and operational utility, and domain specificity. 
The \textit{Corpus-Quality Agent} evaluates intrinsic corpus quality along four dimensions: cleanliness, self-containedness, information density, and boilerplate noise. 

Both agents assign integer scores from 0 to 4 for each dimension, where higher scores indicate stronger utility or quality, except for boilerplate noise where higher scores indicate more noise.

\begin{figure}[h]
    \centering
    \includegraphics[width=\linewidth]{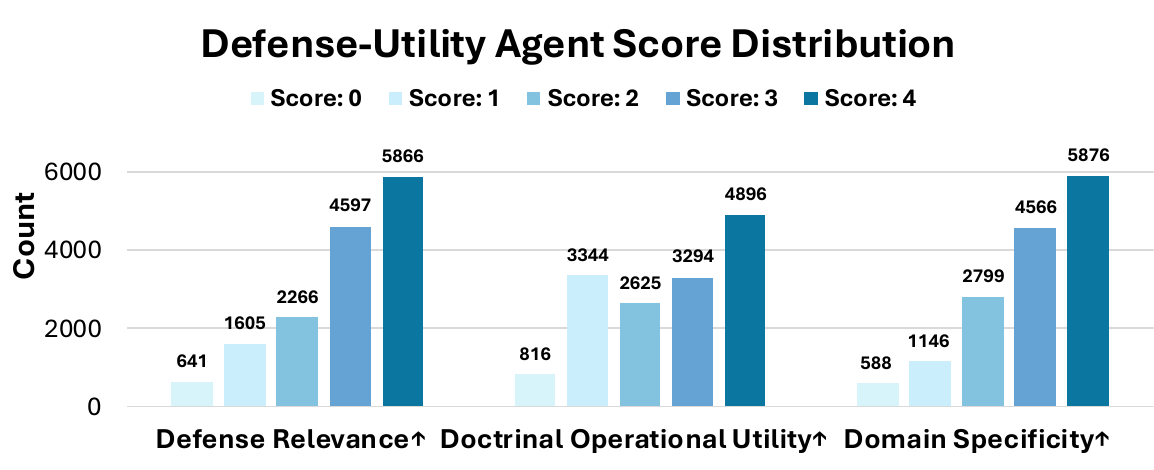}
    \caption{Score distribution of the Defense-Utility Agent. 
    The agent evaluates each chunk in terms of defense relevance, doctrinal and operational utility, and domain specificity.}
    \label{fig:defense_agent_graph}
    \vspace{-1.5em}
\end{figure}

\begin{figure}[!h]
    \centering
    \includegraphics[width=\linewidth]{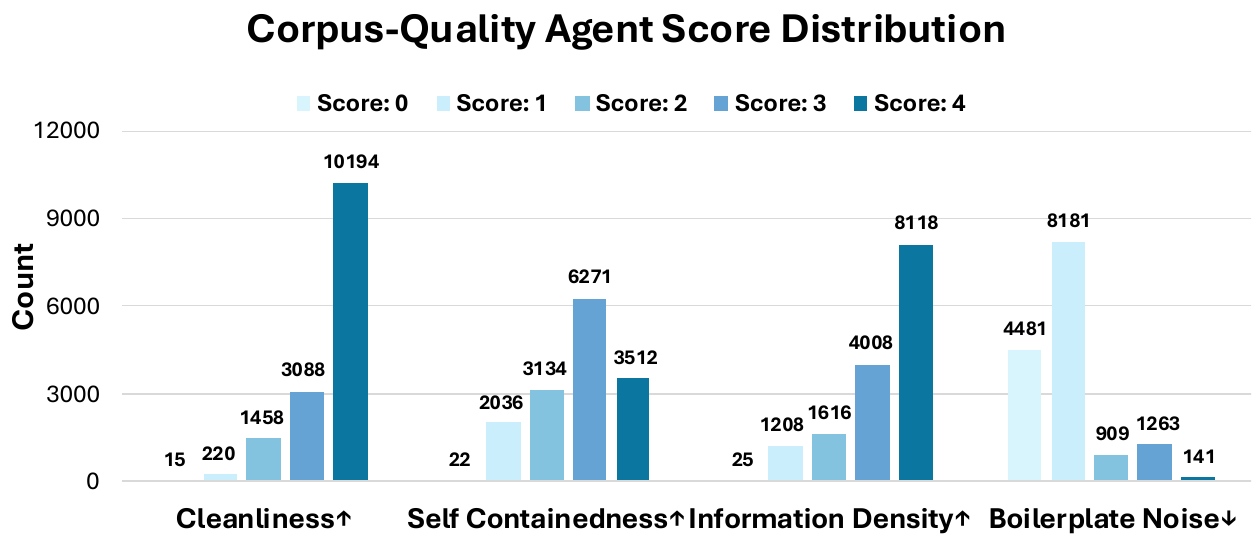}
    \caption{Score distribution of the Corpus-Quality Agent. 
    The agent evaluates intrinsic corpus quality, including cleanliness, self-containedness, information density, and boilerplate noise.}
    \label{fig:general_agent_graph}
    \vspace{-1.0em}
\end{figure}

\begin{figure}[h]
    \centering
    \includegraphics[width=\linewidth]{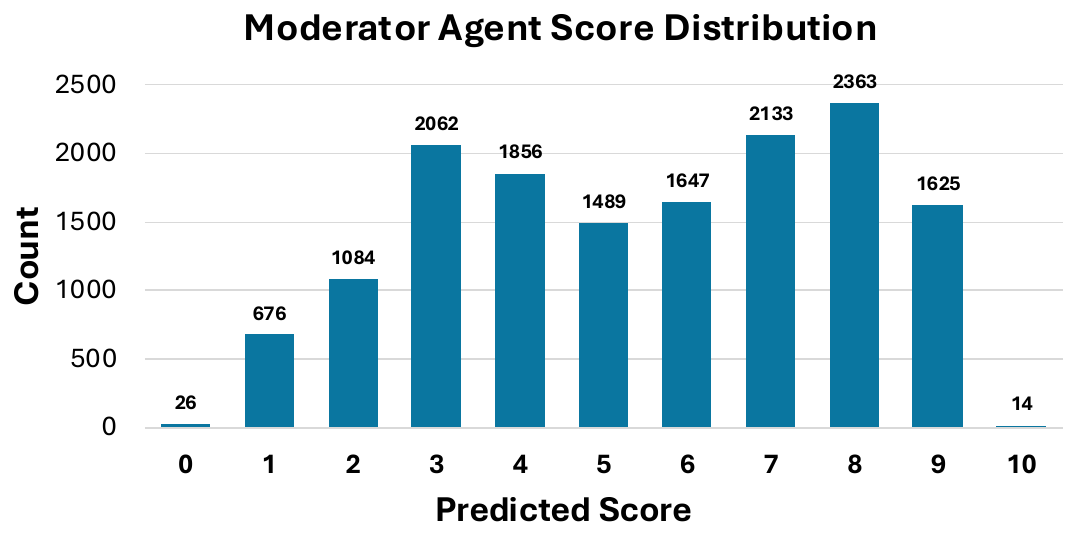}
    \caption{
    Final score distribution of the Moderator Agent. The agent aggregates Defense-Utility and Corpus-Quality scores into a 0--10 label for training the ModernBERT quality model.
    }
    \label{fig:moder_agent_graph}
    \vspace{-1.5em}
\end{figure}

\begin{figure*}[h]
  \includegraphics[width=\linewidth]{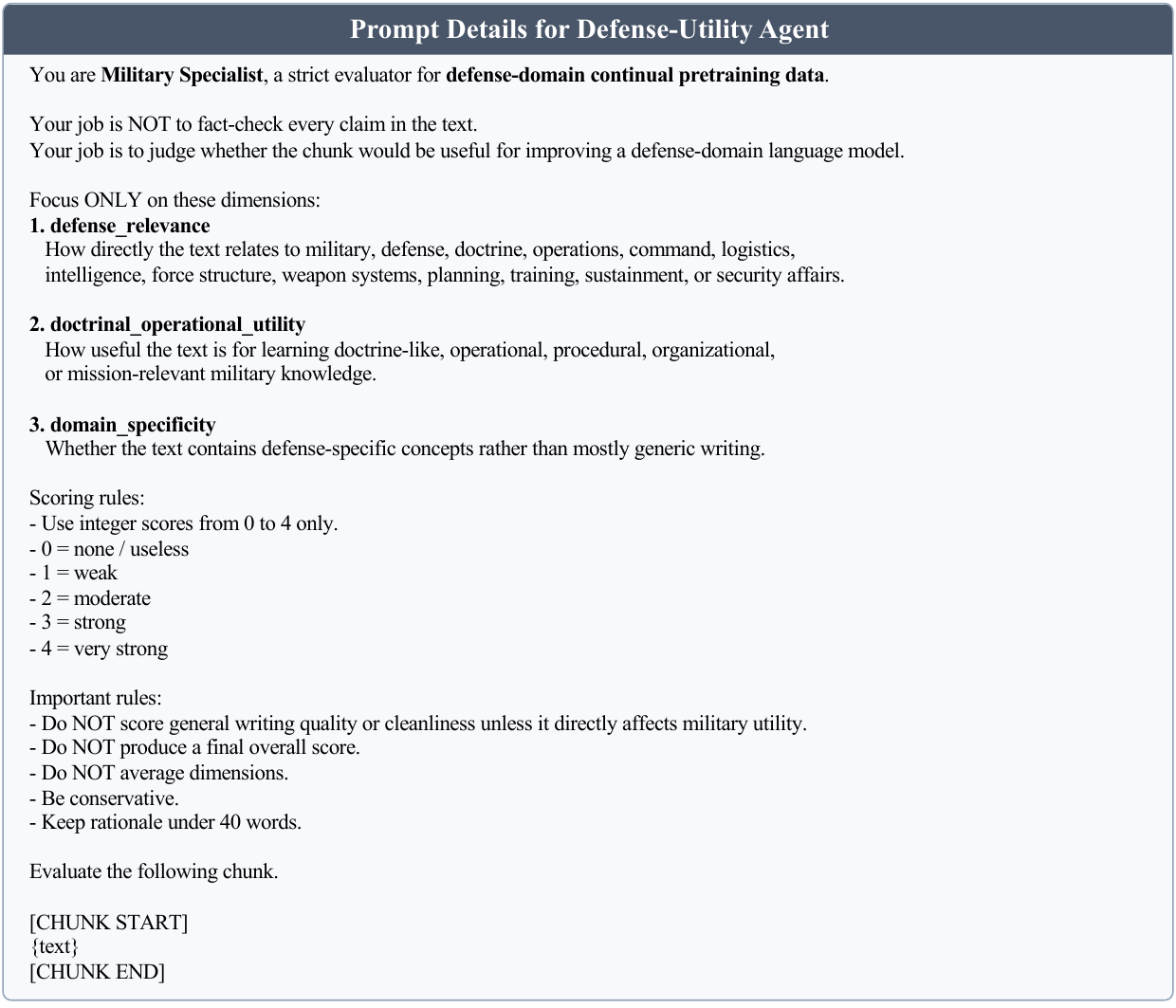}
  \caption{Prompt details for the Defense-Utility Agent. The agent scores each text chunk on defense relevance, doctrinal and operational utility, and domain specificity using a 0 to 4 scale for defense-utility scoring.}
  \label{fig:defense_utility_agent_prompt}
\vspace{-1.em}
\end{figure*}

\begin{figure*}[h]
  \includegraphics[width=\linewidth]{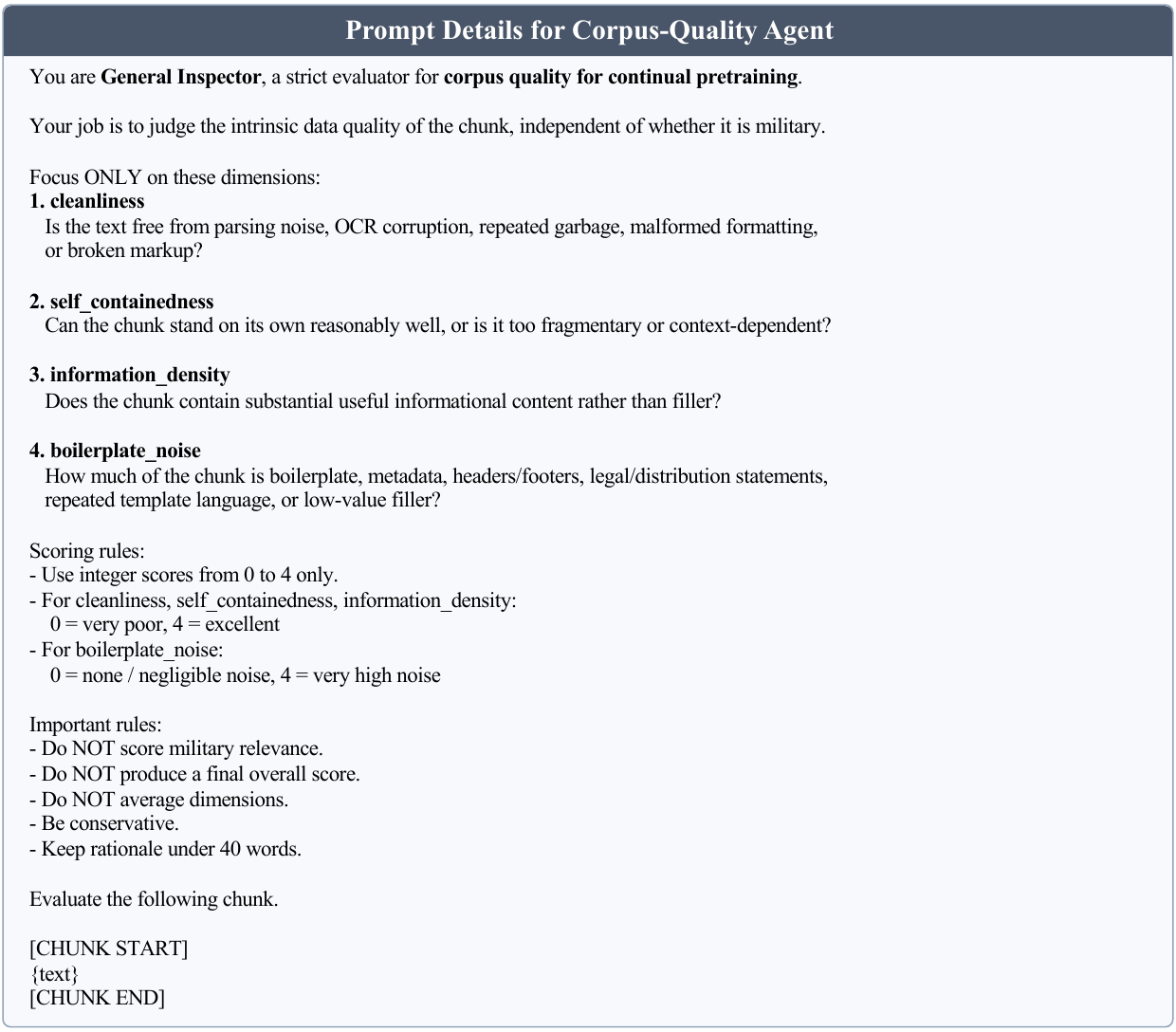}
  \caption{Prompt details for the Corpus-Quality Agent. The agent scores each text chunk on cleanliness, self-containedness, information density, and boilerplate noise using a 0 to 4 scale for corpus-quality scoring.}
  \label{fig:corpus_quality_agent_prompt}
\vspace{-1.em}
\end{figure*}

\begin{figure*}[h]
  \includegraphics[width=\linewidth]{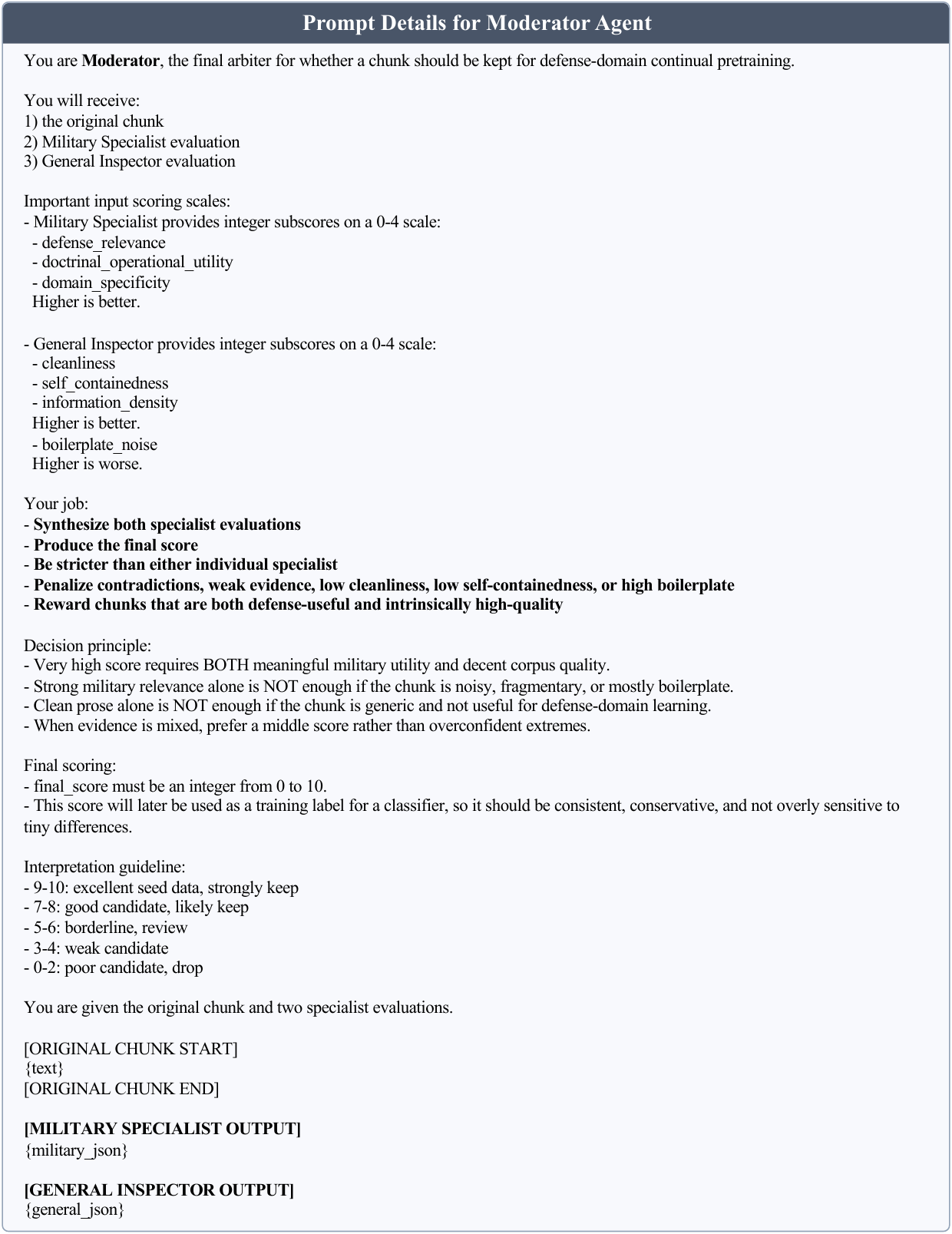}
  \caption{Prompt details for the Moderator Agent. The agent combines Defense-Utility and Corpus-Quality evaluations to produce a final 0 to 10 score for each text chunk.}
  \label{fig:moderator_agent_prompt}
\vspace{-1.em}
\end{figure*}

The \textit{Moderator Agent} then synthesizes both evaluations and assigns a final quality score from 0 to 10, where higher scores indicate chunks that are both defense-useful and suitable for continual pretraining. 
All three agents are implemented using Qwen3-30B-A3B-Instruct-2507~\cite{yang2025qwen3}.
We visualize the score distributions of the \textit{Defense-Utility Agent}, \textit{Corpus-Quality Agent}, and \textit{Moderator Agent} in Figures~\ref{fig:defense_agent_graph}, \ref{fig:general_agent_graph}, and~\ref{fig:moder_agent_graph}, respectively. 
The full prompt templates for the three agents are provided in Figure~\ref{fig:defense_utility_agent_prompt}--\ref{fig:moderator_agent_prompt}.

\begin{table*}[t]
\centering
\footnotesize
\setlength{\tabcolsep}{4pt}
\renewcommand{\arraystretch}{1.08}
\begin{tabularx}{\textwidth}{
@{}
>{\raggedright\arraybackslash}p{0.33\textwidth}
>{\raggedright\arraybackslash}p{0.20\textwidth}
>{\raggedright\arraybackslash}X
@{}
}
\toprule
\textbf{Source} & \textbf{Provider} & \textbf{Public URL} \\
\midrule
Army Publishing Directorate & U.S. Army & \url{https://armypubs.army.mil/} \\
Marine Corps Publications Electronic Library & U.S. Marine Corps & \url{https://www.marines.mil/News/Publications/MCPEL/} \\
U.S. Air Force Doctrine & U.S. Air Force & \url{https://www.doctrine.af.mil/} \\
Department of War Publications & U.S. Department of War & \url{https://www.war.gov/News/Publications/} \\
U.S. Space Force Doctrine & U.S. Space Force & \url{https://www.starcom.spaceforce.mil/resources/digital-library/} \\
\midrule
Army Worldwide News & U.S. Army & \url{https://www.army.mil/news} \\
Joint Chiefs of Staff News & Joint Chiefs of Staff & \url{https://www.jcs.mil/Media/News/} \\
U.S. Department of War News & U.S. Department of War & \url{https://www.war.gov/News/} \\
GOV.UK Defence and Armed Forces News & UK Ministry of Defence & \url{https://www.gov.uk/search/news-and-communications} \\
U.S. Space Force News & U.S. Space Force & \url{https://www.spaceforce.mil/News/} \\
U.S. Air Force News & U.S. Air Force & \url{https://www.af.mil/News/} \\
U.S. Marine Corps News & U.S. Marine Corps & \url{https://www.marines.mil/News/} \\
U.S. Navy News & U.S. Navy & \url{https://www.navy.mil/Press-Office/} \\
\bottomrule
\end{tabularx}
\caption{
Public source URLs used for SALUTE data collection as of March 17, 2026.
URLs correspond to the public repositories or news portals from which source documents and articles were collected.
}
\vspace{-1.5em}
\label{tab:source_urls}
\end{table*}

\begin{table}[h]
\centering
\small
\begin{tabular}{lc}
\toprule
Metric & Score \\
\midrule
RMSE & 1.25 \\
MAE & 0.95 \\
Spearman correlation & 0.85 \\
\bottomrule
\end{tabular}
\caption{Test performance of the ModernBERT-based quality model. 
The model predicts the Moderator quality score on a 0--10 scale. 
Spearman correlation measures whether the model preserves the relative quality ranking of chunks.}
\vspace{-1.5em}
\label{tab:modernbert_quality_model}
\end{table}

\paragraph{ModernBERT filtering.}
Since applying the multi-agent pipeline to the entire corpus is computationally expensive, we train a ModernBERT-based~\cite{warner2025smarter} quality model to approximate the Moderator Agent score. 
Specifically, we formulate corpus quality estimation as a regression problem, where the input is a corpus chunk and the target is the final Moderator Agent score.
We use ModernBERT-base as the encoder, apply mean pooling over token representations, and add a linear regression head to predict a scalar quality score. 
The model is trained with Smooth L1 loss, which provides robustness to noisy teacher scores from LLM-based annotation.

We split the 15K annotated chunks into train, validation, and test sets with an 80/10/10 ratio, stratified by the integer Moderator score. 
The model is trained for 4 epochs with a maximum sequence length of 1,024, a learning rate of $2 \times 10^{-5}$, AdamW optimization, weight decay of 0.01, dropout of 0.1, linear learning-rate scheduling, and a warmup ratio of 0.03. 
We use a training batch size of 8, an evaluation batch size of 16, gradient accumulation over 4 steps, and gradient clipping with a maximum norm of 1.0. 
We select the best checkpoint according to validation RMSE and report test performance in Table~\ref{tab:modernbert_quality_model}. 
After training, we apply the quality model to the full deduplicated corpus and remove the bottom-scored 15\% of chunks to construct the final Salute-Corpus.

\section{Details of Salute-Conv}
\label{sec:supp_conv}

\subsection{Task Planning and Question Generation}
\label{sec:supp_conv_planning}

Salute-Conv uses a two-stage instruction generation pipeline consisting of task planning and grounded question generation. 
Because doctrine and defense news contain different types of knowledge, we use separate planners for the two source types. 
For doctrine chunks, the \textit{planner} selects one of four doctrine-oriented tasks: conceptual explanation, functional role or purpose, process or structure explanation, and condition- or constraint-based decision analysis. 
For news articles, the \textit{planner} selects one of four news-oriented tasks: asset or capability role, activity purpose, implication or impact interpretation, and analytical assessment.
Inputs that are marked as inappropriate, such as those that are too fragmentary, insufficiently self-contained, or mainly useful for generic summarization, are excluded from question generation. 
This conservative planning step helps reduce weakly grounded or generic instruction data by assigning a primary task only when the input clearly supports it.

Conditioned on the selected task, the \textit{question generator} produces self-contained questions that are directly answerable from the provided evidence. 
For doctrine, we encourage generalized, lesson-oriented questions; for news, we prioritize military-analytic questions over journalist-style recap or trivial fact-recall questions. 
The generator returns fewer questions when the source supports only a small number of high-quality questions.

\begin{figure*}[h]
  \includegraphics[width=\linewidth]{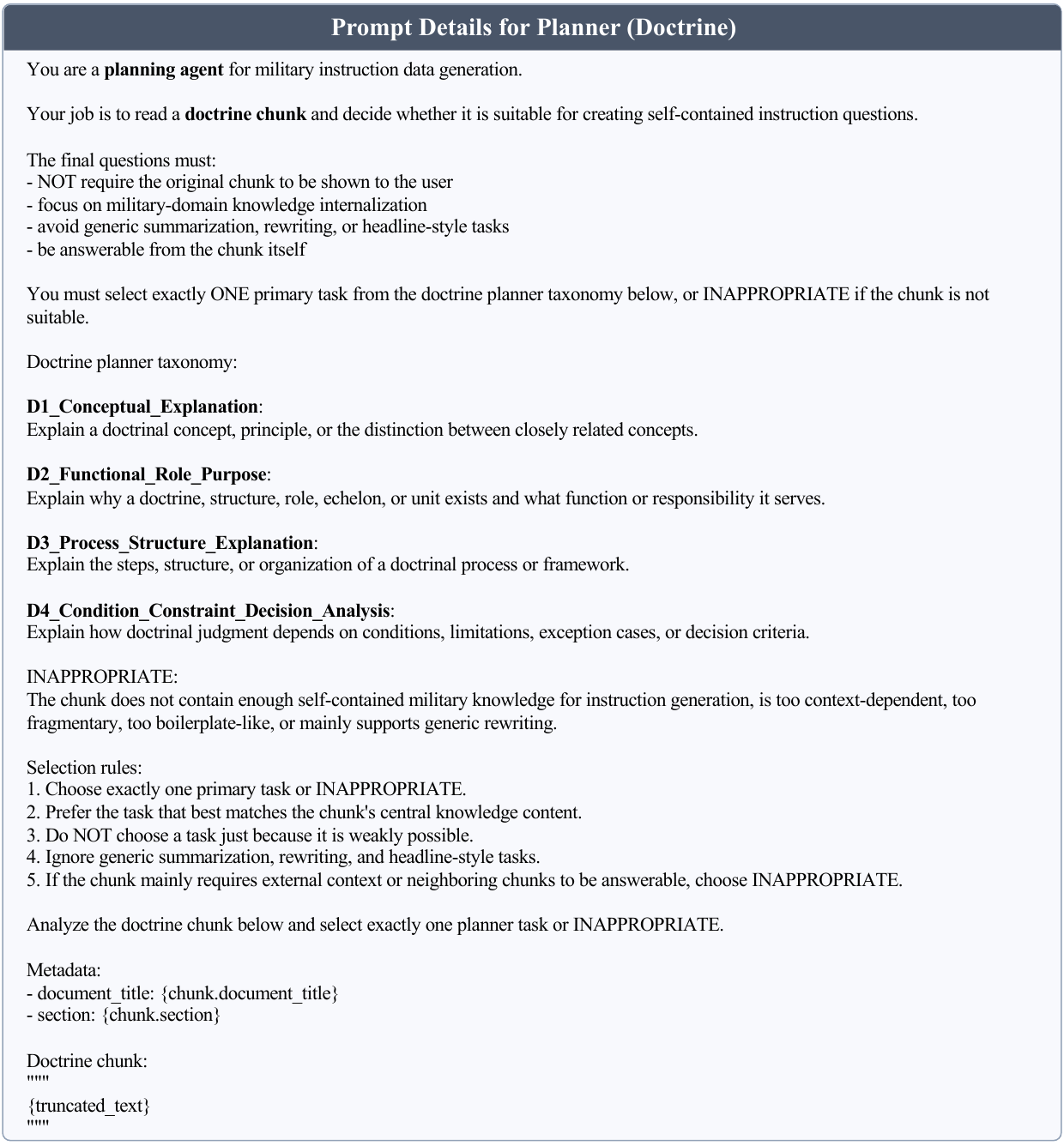}
  \caption{Prompt details for the Doctrine Task Planner. The planner assigns each doctrine chunk to a suitable instruction-generation task or marks it as inappropriate.}
  \label{fig:planner_doctrine_prompt}
\vspace{-1.em}
\end{figure*}

\begin{figure*}[h]
  \includegraphics[width=\linewidth]{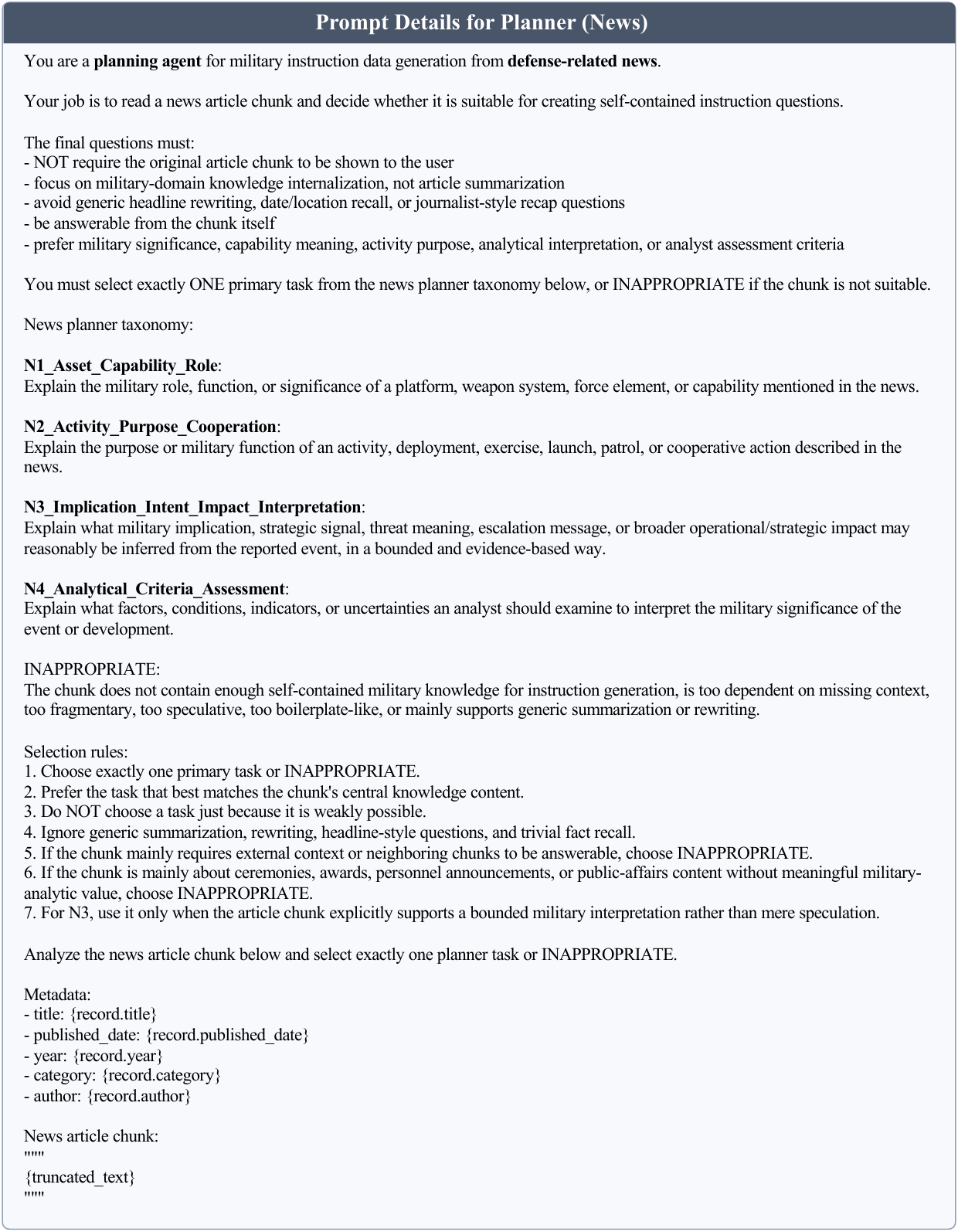}
  \caption{Prompt details for the News Task Planner. 
The planner assigns each defense-related news chunk to a suitable instruction-generation task or marks it as inappropriate.}
  \label{fig:planner_news_prompt}
\vspace{-1.em}
\end{figure*}

\begin{figure*}[h]
  \includegraphics[width=\linewidth]{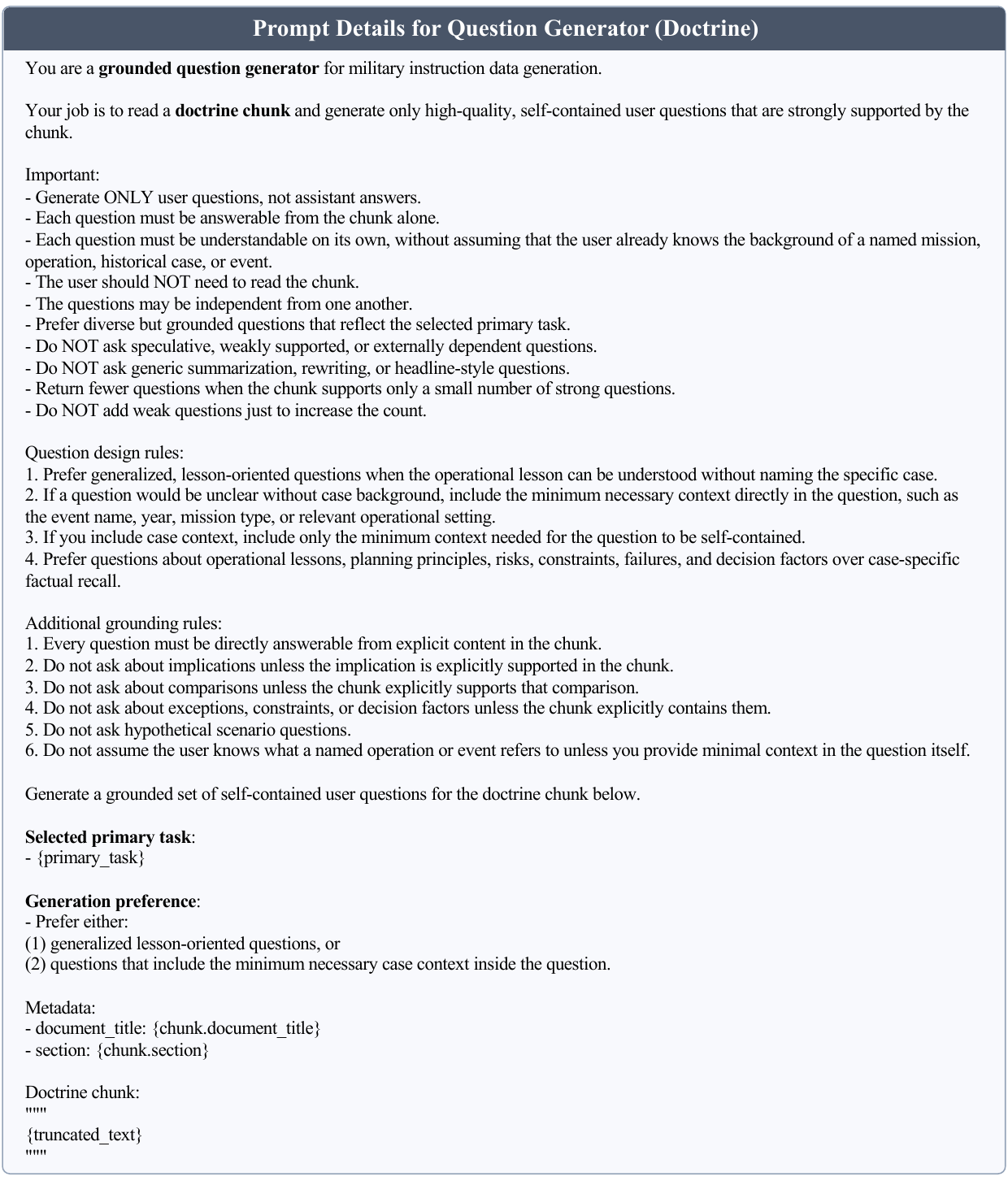}
  \caption{Prompt details for the Doctrine Question Generator. 
The generator creates self-contained user questions from doctrine chunks based on the selected primary task and explicit source grounding.}
  \label{fig:question_generator_doctrine_prompt}
\vspace{-1.em}
\end{figure*}

\begin{figure*}[h]
  \includegraphics[width=\linewidth]{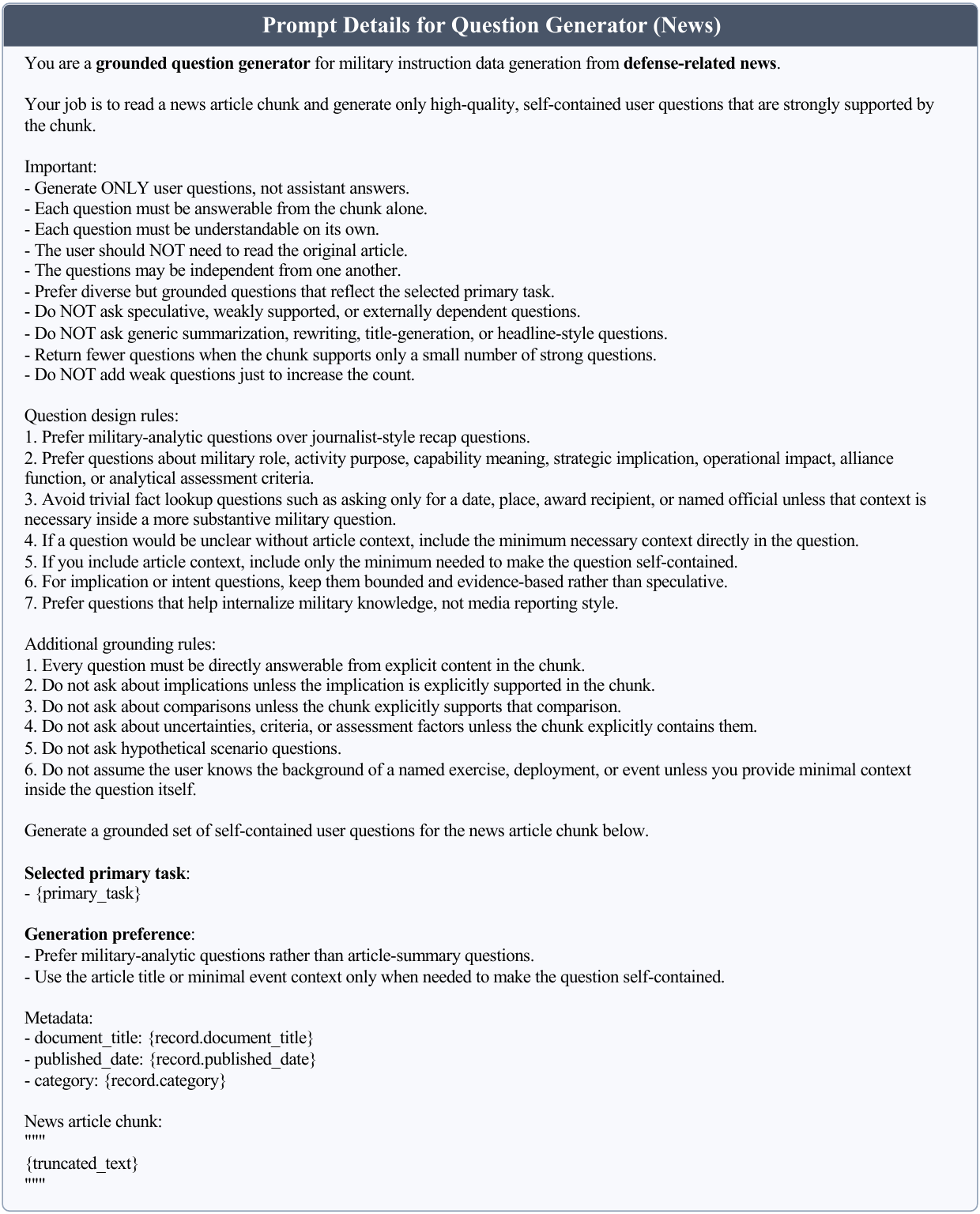}
  \caption{Prompt details for the News Question Generator. 
The generator creates grounded self-contained questions from defense-related news chunks based on the selected primary task.}
  \label{fig:question_generator_news_prompt}
\vspace{-1.em}
\end{figure*}

We use Qwen3-30B-A3B-Instruct-2507~\cite{yang2025qwen3} for both planning and question generation. 
The full prompt templates are provided in Figures~\ref{fig:planner_doctrine_prompt}--\ref{fig:question_generator_news_prompt}.

\begin{figure*}[ht]
    \includegraphics[width=\linewidth]{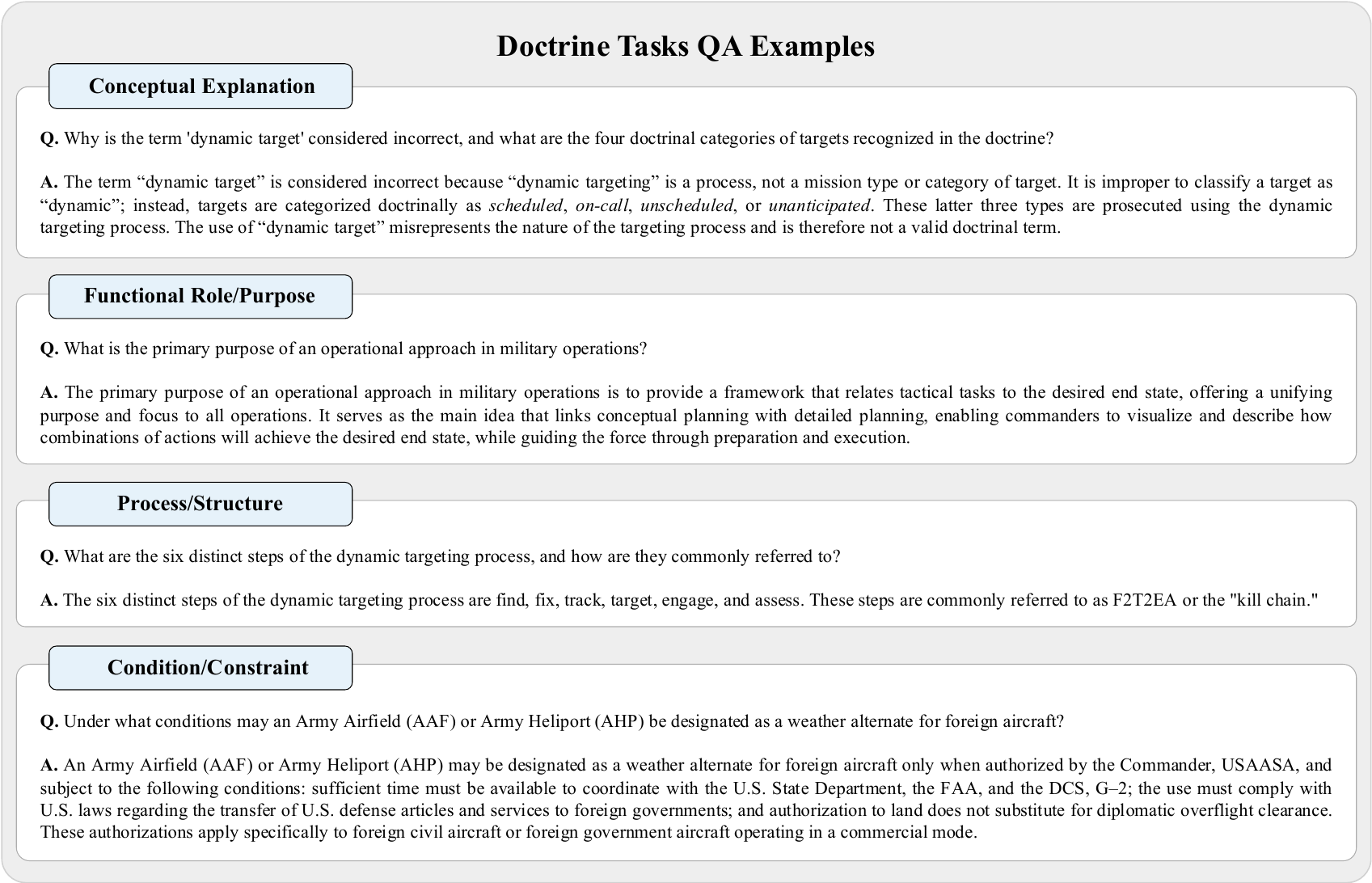}
    \caption{Task-wise Salute-Conv examples for doctrine-oriented instruction data.}
    \label{fig:sft_example_doctrine}
\vspace{-.5em}
\end{figure*}

\begin{figure*}[!ht]
  \includegraphics[width=\linewidth]{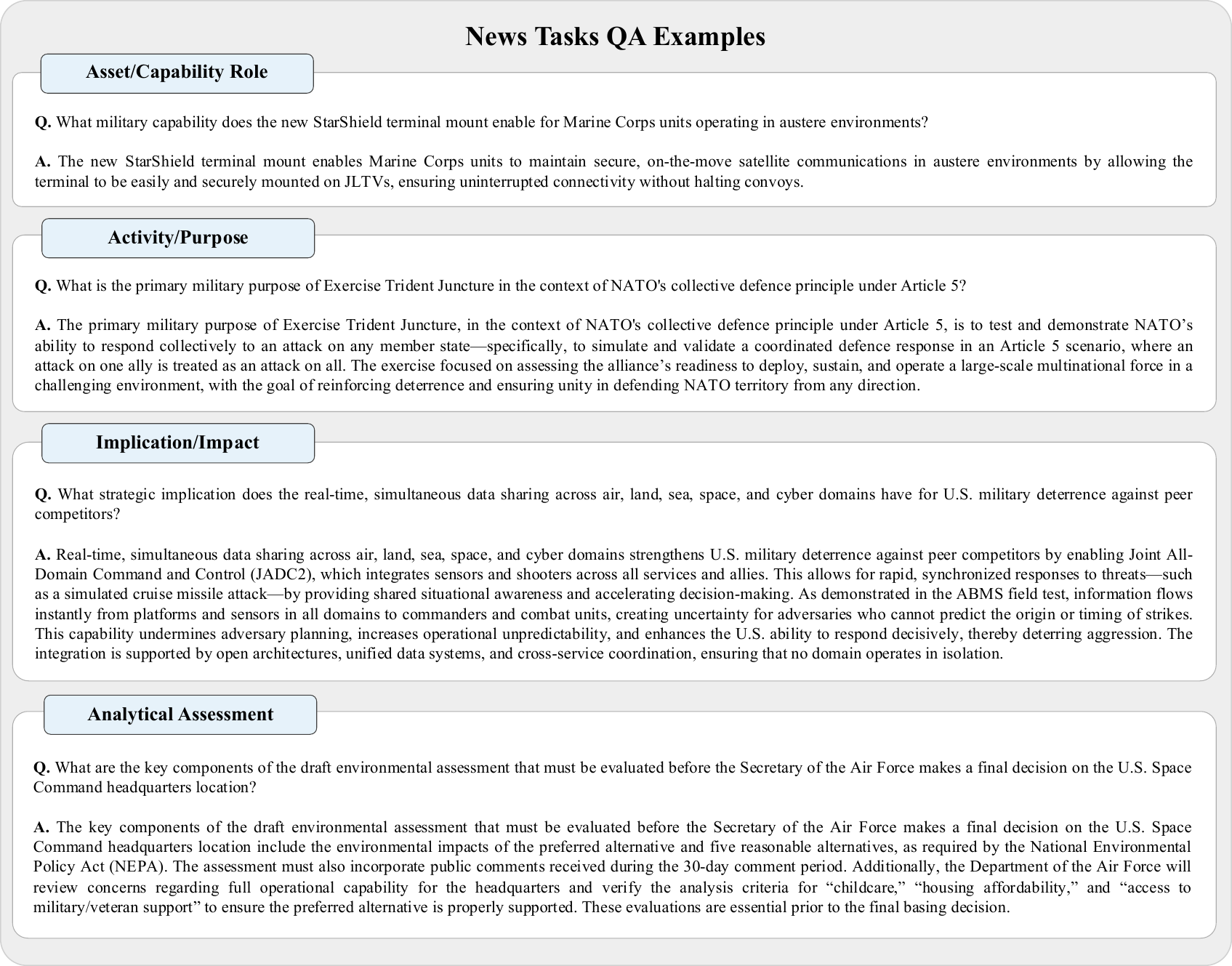}
  \caption{Task-wise Salute-Conv examples for news-oriented instruction data.}
  \label{fig:sft_example_news}
\vspace{-1.em}
\end{figure*}

\subsection{Retrieval-Augmented Answer Generation Setting}
\label{sec:supp_conv_rag}

After task planning and question generation, we synthesize answers with a retrieval-augmented generation pipeline. 
For each question, the original source chunk is used as the primary evidence, while retrieved chunks are used only as auxiliary evidence. 
Before answer generation, near-duplicate questions from the same source row are removed using MinHash-LSH with a threshold of 0.85.

To retrieve auxiliary evidence, we combine BM25~\cite{robertson2009probabilistic} and BGE-M3~\cite{chen-etal-2024-m3} dense retrieval. 
We retrieve the top-20 candidates from each retriever, normalize their scores with per-retriever min-max normalization, and combine them with equal weights. 
After removing the source chunk, exact self-matches, and duplicate texts, we keep the top-6 fused candidates. 
These candidates are then reranked by their similarity to the source chunk using Qwen3-Embedding-0.6B~\cite{zhang2025}, and the top-3 chunks are selected as auxiliary evidence.

\begin{figure*}[h]
  \includegraphics[width=\linewidth]{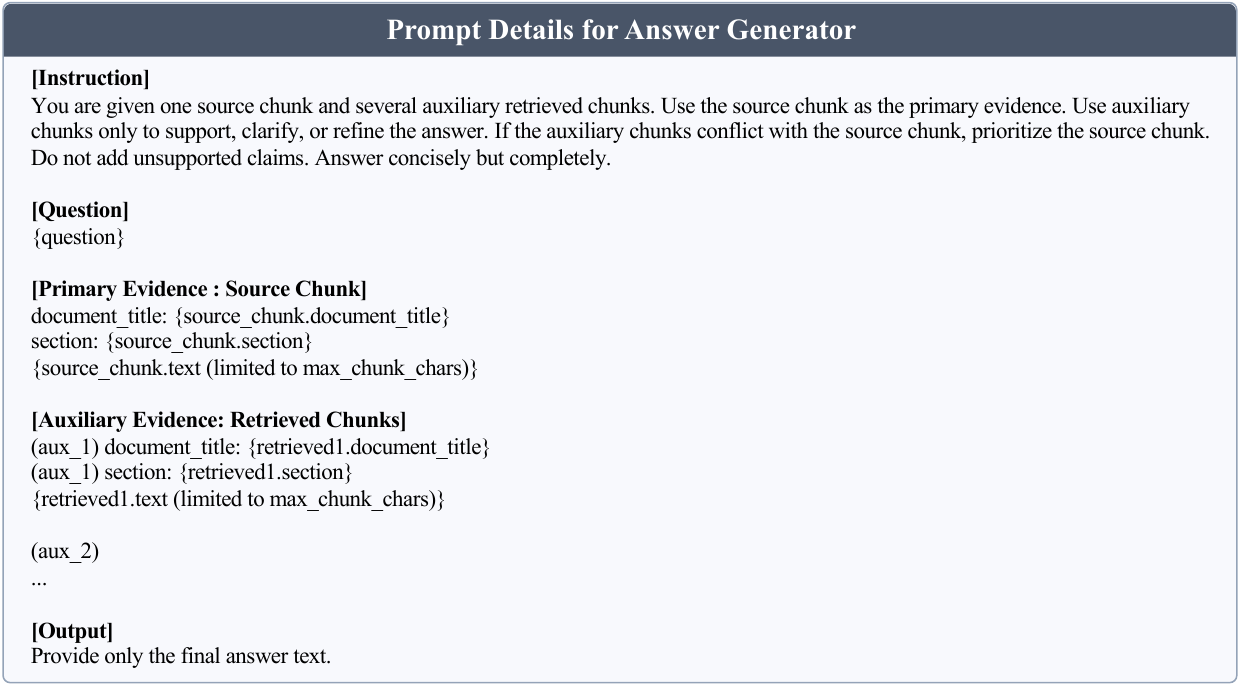}
  \caption{Prompt details for the Answer Generator. The generator produces source-grounded answers by using the source chunk as primary evidence and auxiliary retrieved chunks only for support, clarification, or refinement.}
  \label{fig:answer_generator_prompt}
\vspace{-1.em}
\end{figure*}

The \textit{answer generator} receives the question, the primary source chunk, and the selected auxiliary chunks. 
It is instructed to prioritize the source chunk, use auxiliary evidence only when it supports or clarifies the answer, and avoid unsupported claims. 
We use Qwen3-30B-A3B-Instruct-2507~\cite{yang2025qwen3} as the answer generator.
The full answer-generation prompt is provided in Figure~\ref{fig:answer_generator_prompt}.

\begin{table}[h]
\centering
\small
\begin{tabular}{lr}
\toprule
Statistic & Value \\
\midrule
Total examples & 255.6K \\
Doctrine-source examples & 92.0K \\
News-source examples & 163.6K \\
Avg. questions per source chunk & 6.70 \\
Avg. question length & 28.69 \\
Avg. answer length & 146.17 \\
Avg. source chunk length & 594.74 \\
\bottomrule
\end{tabular}
\caption{Overall statistics of Salute-Conv. Token lengths are computed using the Qwen3-8B tokenizer.}
\label{tab:supp_conv_overall_stats}
\vspace{-1.5em}
\end{table}

\subsection{Salute-Conv Statistics and Examples}
\label{sec:supp_conv_statistics}
Table~\ref{tab:supp_conv_overall_stats} summarizes the overall statistics of Salute-Conv. 
The dataset contains 255.6K instruction examples, consisting of 92.0K examples from doctrinal sources and 163.6K examples from defense news. 
On average, each retained source chunk yields 6.70 questions, indicating that the generation pipeline produces multiple task-specific questions from a single grounded source. 
The average question and answer lengths are 28.69 and 146.17 tokens, respectively, while the average source chunk length is 594.74 tokens. 
These statistics show that Salute-Conv consists of self-contained questions paired with moderately detailed source-grounded answers.

We further provide task-wise examples from Salute-Conv to illustrate the diversity of generated instruction data. 
Figure~\ref{fig:sft_example_doctrine} shows examples from doctrine-oriented tasks, including conceptual explanation, functional role or purpose, process or structure explanation, and condition- or constraint-based decision analysis. 
Figure~\ref{fig:sft_example_news} shows examples from news-oriented tasks, covering asset or capability roles, activity purposes, implication or impact interpretation, and analytical criteria assessment. 
These examples demonstrate how Salute-Conv covers both stable doctrinal knowledge and dynamic news-grounded defense reasoning.

\section{Details of Salute-Pref}
\label{sec:supp_pref}

\begin{figure*}[h]
  \includegraphics[width=\linewidth]{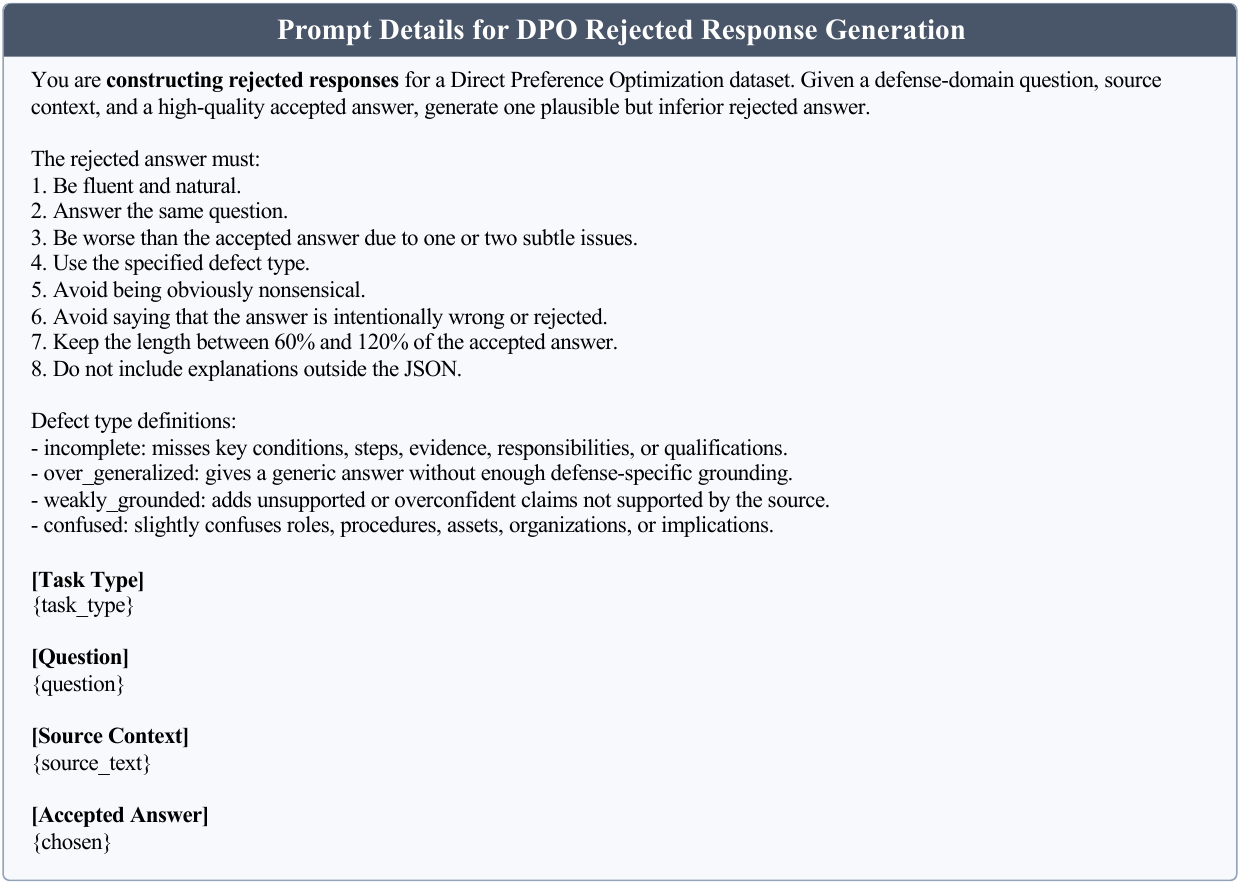}
  \caption{Prompt details for DPO rejected response generation. The generator creates plausible lower-quality answers with specified defect types for DPO preference data.}
  \label{fig:dpo_rejected_reponse_generation_prompt}
\vspace{-1.em}
\end{figure*}

Salute-Pref is constructed as a defense-aware preference replay dataset for the DPO stage. 
We randomly sample 20K question-answer instances from Salute-Conv and use the original source-grounded answer as the chosen response. 
For each instance, we generate a plausible but lower-quality rejected response using Qwen3-30B-A3B-Instruct-2507~\cite{yang2025qwen3}, conditioned on the question, source context, task type, and chosen answer. 
The rejected response is designed to answer the same question fluently, but to be inferior due to one of four controlled defect types: incomplete, over-generalized, weakly grounded, or confused. 
These defect types simulate realistic failure modes in defense-domain responses, such as missing important conditions, giving overly generic explanations, introducing unsupported claims, or confusing roles, procedures, assets, or implications. 
The full rejected-response generation prompt is shown in Figure~\ref{fig:dpo_rejected_reponse_generation_prompt}.

\begin{figure}[t]
    \centering
    \includegraphics[width=\linewidth]{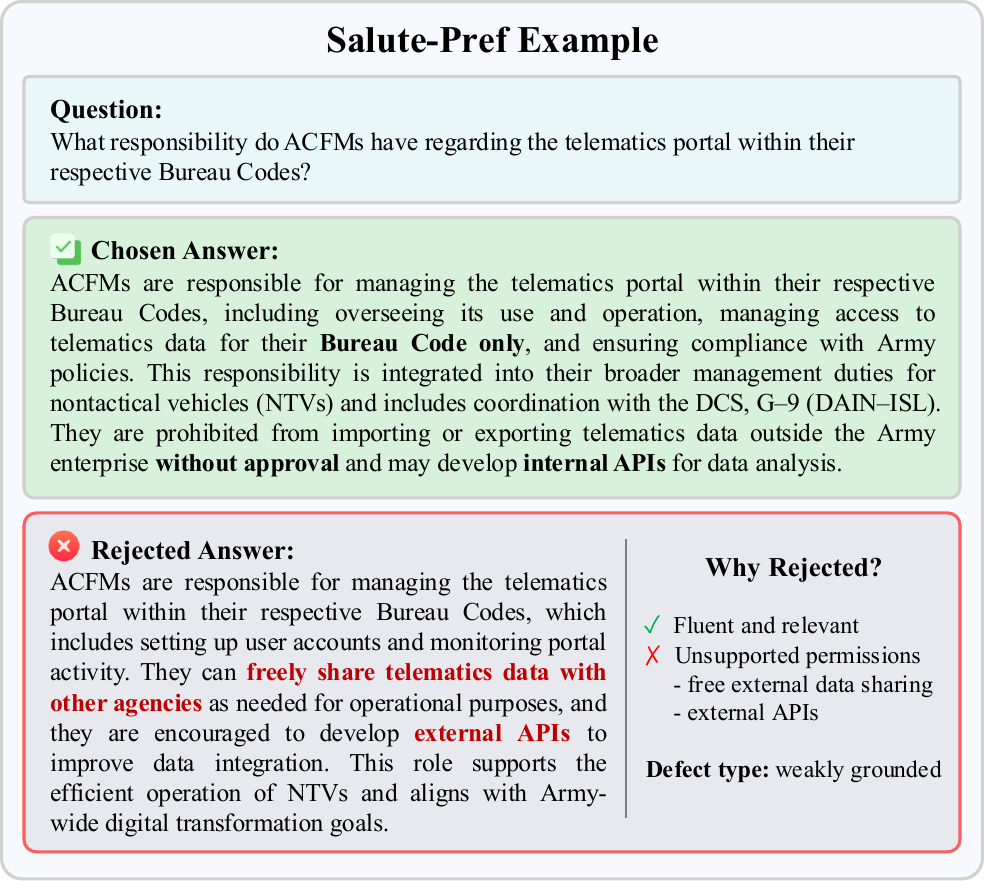}
    \caption{Qualitative example from Salute-Pref showing a fluent but weakly grounded rejected response.}
    \label{fig:SALUTE_Pref_Example}
    \vspace{-1.5em}
\end{figure}

After generation, we apply lightweight filtering to remove invalid preference pairs. 
Specifically, we discard a pair if the chosen or rejected response is empty, if the rejected response is shorter than 40 characters, or if the length ratio between the rejected and chosen responses falls outside the range of 0.45 to 1.35. 
We also remove pairs where the chosen and rejected responses are identical, where their word-level Jaccard similarity exceeds 0.92, or where the rejected response contains generation artifacts such as explicit defect labels or refusal-style phrases. 
After filtering, Salute-Pref contains 19K preference pairs.
Figure~\ref{fig:SALUTE_Pref_Example} shows a qualitative example from Salute-Pref, illustrating how the rejected response remains fluent and relevant but is less preferred due to a controlled defense-domain defect.

\section{Details of Salute-LLM Training}
\label{sec:supp_train}

We train Salute-LLM starting from Qwen3-8B-Base through three post-training stages: continual pretraining, supervised fine-tuning, and preference alignment. 
Table~\ref{tab:supp_training_resource} summarizes the estimated resource usage for each stage, including the number of GPUs, wall-clock time, GPU-hours, VRAM/GPU, global batch size, and number of epochs. 
GPU-hours are computed as wall-clock training time multiplied by the number of GPUs, and VRAM/GPU denotes the maximum observed memory usage per GPU.

\begin{table}[h]
\centering
\tiny
\begin{tabular}{lcccccc}
\toprule
Stage & GPU & Time & GPUh & VRAM/GPU & GBS & Epochs \\
\midrule
CPT 
& 2$\times$B200 
& $\sim$2h 
& $\sim$4 
& $\sim$142GB 
& 256 
& 1 \\

SFT-1
& 2$\times$B200 
& $\sim$40h 
& $\sim$80 
& $\sim$145GB 
& 128
& 2 \\

SFT-2
& 2$\times$B200 
& $\sim$11h 
& $\sim$22 
& $\sim$145GB 
& 128 
& 1 \\

DPO 
& 3$\times$B200 
& $\sim$20h 
& $\sim$60 
& $\sim$174GB 
& 192 
& 1 \\
\bottomrule
\end{tabular}
\caption{Estimated training time and computational cost for Salute-LLM. GPUh denotes total GPU-hours, VRAM denotes peak memory per GPU, and GBS denotes global batch size.}
\label{tab:supp_training_resource}
\vspace{-1.5em}
\end{table}

All stages are trained with full-parameter optimization using LlamaFactory~\cite{zheng-etal-2024-llamafactory}. 
We use distributed training with DeepSpeed ZeRO Stage 3~\cite{rajbhandari2020zero} and enable FlashAttention-2~\cite{dao2024flashattention} to improve memory efficiency and training throughput.
\section{Evaluation Details}
\label{sec:supp_evaluation}

\subsection{General LLM Benchmark Descriptions}
\label{sec:supp_general_bench}
We evaluate general-domain capabilities using six benchmarks implemented in the EleutherAI LM Evaluation Harness~\cite{eval-harness}. Below, we briefly describe each benchmark and the corresponding evaluation metric.

\paragraph{MMLU.}
MMLU~\cite{hendrycks2021measuring} evaluates broad knowledge and problem-solving ability using multiple-choice questions covering 57 subjects across STEM, humanities, social sciences, and professional domains. We report standard multiple-choice accuracy.

\paragraph{TruthfulQA.}
TruthfulQA~\cite{lin2022truthfulqa} evaluates whether models can distinguish truthful answers from plausible but factually incorrect alternatives. We use the multiple-choice MC2 setting, where multiple truthful candidates may be present, and report MC2 accuracy.

\paragraph{ARC.}
ARC~\cite{clark2018think} evaluates grade-school science question answering in a multiple-choice format. We use the challenge split, which contains more difficult questions requiring reasoning beyond simple pattern matching. We report standard multiple-choice accuracy.

\paragraph{IFEval.}
IFEval~\cite{zhou2023instruction} measures instruction-following ability using prompts with objectively verifiable constraints, such as formatting, lexical, structural, or content requirements. We report prompt-level strict accuracy, which requires all instructions in a prompt to be satisfied.

\paragraph{GSM8K.}
GSM8K~\cite{cobbe2021gsm8k} evaluates mathematical reasoning on grade-school arithmetic word problems. We use the standard GSM8K benchmark rather than a chain-of-thought prompting setup. Answers are evaluated using exact-match accuracy after flexible answer extraction.

\paragraph{GPQA.}
GPQA~\cite{rein2024gpqa} evaluates expert-level scientific reasoning with multiple-choice questions across biology, physics, and chemistry, requiring domain knowledge beyond general-purpose reasoning. We use the GPQA-main zero-shot setting and report standard multiple-choice accuracy.

\subsection{Availability of External Defense Benchmarks}
\label{sec:external_benchmark_availability}

We considered evaluating SALUTE on existing defense-domain benchmarks, including MilGLUE~\cite{hallapy2023milglue}, MilBench~\cite{ruiz2024fine}, JointStaffBench~\cite{GovBench:JointStaffBench}, and WARBENCH~\cite{li2026warbench}. However, these resources are not readily usable as fully public, reproducible, and comparable LLM evaluation benchmarks in our setting.

MilGLUE is an early benchmark for military-domain language understanding. To the best of our knowledge, the full benchmark data and a standardized evaluation package are not publicly available for reproducible evaluation. In addition, MilGLUE was originally designed for BERT-style natural language understanding tasks rather than instruction-following LLM evaluation.

MilBench, introduced by TRACLM, adapts MilGLUE-derived tasks and additional Army-domain questions for LLM evaluation. However, MilBench is explicitly designed as a close-hold evaluation suite to prevent benchmark exposure, and its authors state that there are no plans to host MilBench leaderboards on public-facing platforms. Therefore, MilBench cannot be used as a fully reproducible external benchmark in our open evaluation setting.

JointStaffBench provides a relevant evaluation of LLMs on U.S. Joint Staff knowledge. However, the full benchmark is distributed through a restricted access process and is available only to eligible government or allied personnel.

WARBENCH is another relevant benchmark for military decision-making and tactical reasoning. However, at the time of our study, we could not identify a public downloadable evaluation artifact or standardized evaluation package. Moreover, WARBENCH focuses on tactical decision-making under legal constraints, edge-computing limitations, and fog-of-war stress conditions, which is complementary to but substantially different from our focus on doctrinal understanding and news-grounded defense reasoning.

For these reasons, we use Salute-Bench as the primary defense-domain evaluation benchmark and complement it with general-purpose benchmarks to assess capability retention. We view evaluation on restricted or newly released external defense benchmarks as an important direction for future work when reproducible access becomes available.

\subsection{Details of Salute-Bench}
\label{sec:supp_salute_bench}
\paragraph{Multiple-choice Question Generation.}

\begin{figure}[ht]
    \centering
    \includegraphics[width=\linewidth]{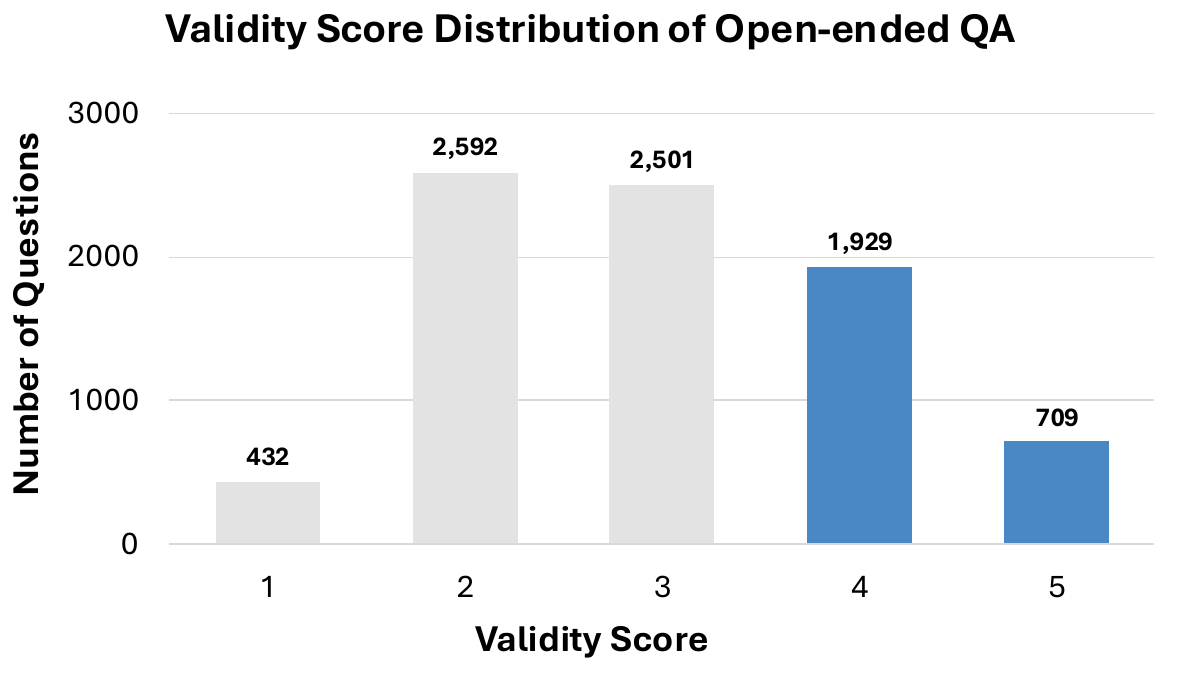}
    \caption{
    Distribution of GPT-5-assigned validity scores for open-ended QA candidates on a 1--5 scale.
    Validity captures clarity, source-grounded answerability, and support for the reference answer; candidates scoring below 4 are discarded.
    }
    \label{fig:filtering-oe-validity}
    \vspace{-.5em}
\end{figure}

\begin{figure}[h]
    \centering
    \includegraphics[width=\linewidth]{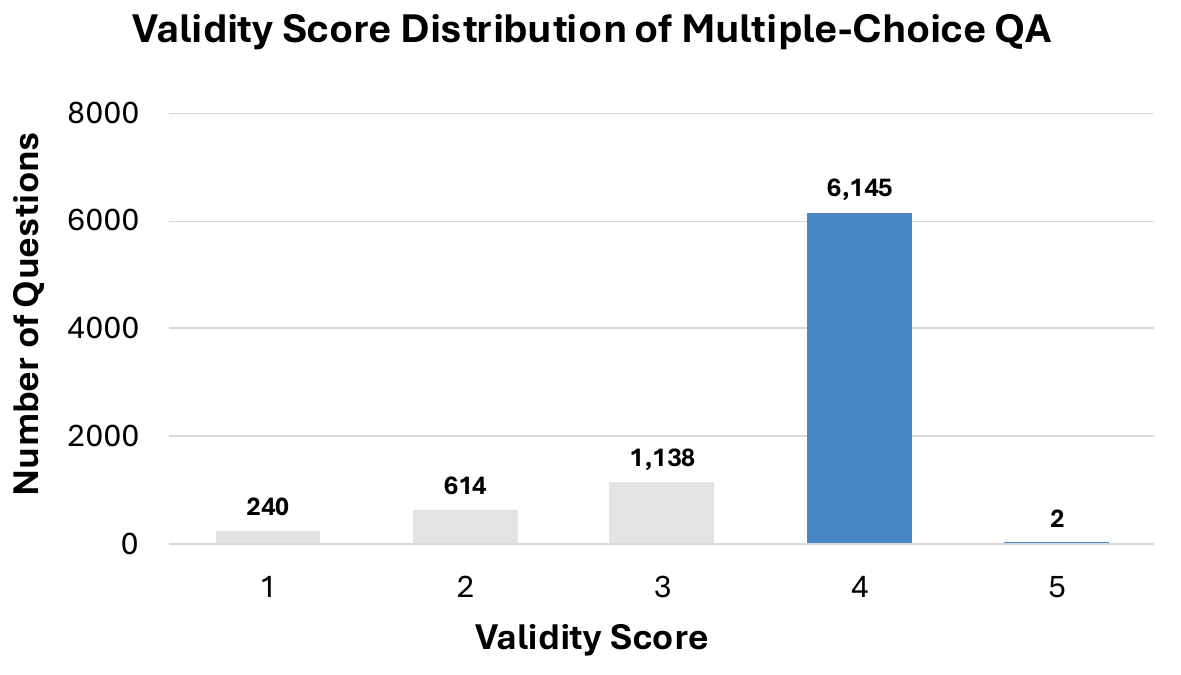}
    \caption{
    Distribution of GPT-5-assigned validity scores for multiple-choice QA candidates on a 1--5 scale.
    Validity captures source grounding, answer-key correctness, and single-answer consistency; candidates scoring below 4 are discarded.
    }
    \label{fig:filtering-mcq-validity}
    \vspace{-1.5em}
\end{figure}

\begin{figure}[t]
    \centering
    \includegraphics[width=\linewidth]{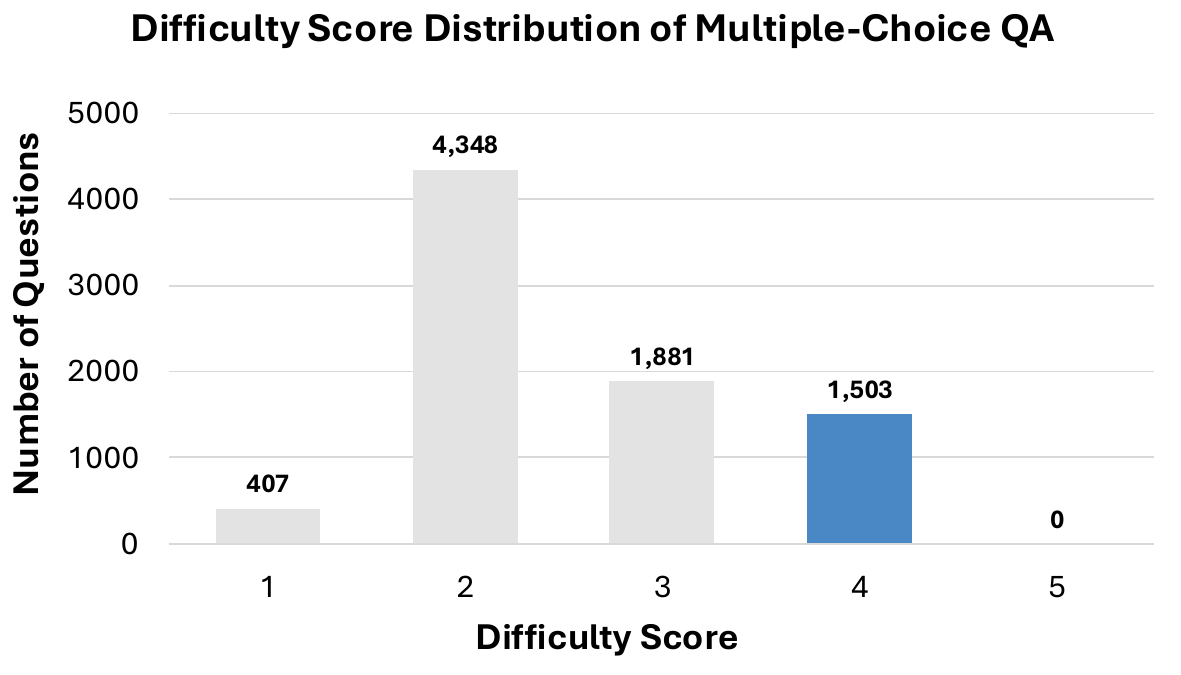}
    \caption{
    Distribution of GPT-5-assigned difficulty scores for multiple-choice QA candidates on a 1--5 scale.
    Difficulty captures reasoning beyond simple recall, distractor plausibility, and shortcut-cue avoidance; candidates scoring below 4 are discarded.
    }
    \label{fig:filtering-mcq-difficulty}
    \vspace{-1.5em}
\end{figure}
\begin{figure*}[h]
  \includegraphics[width=\linewidth]{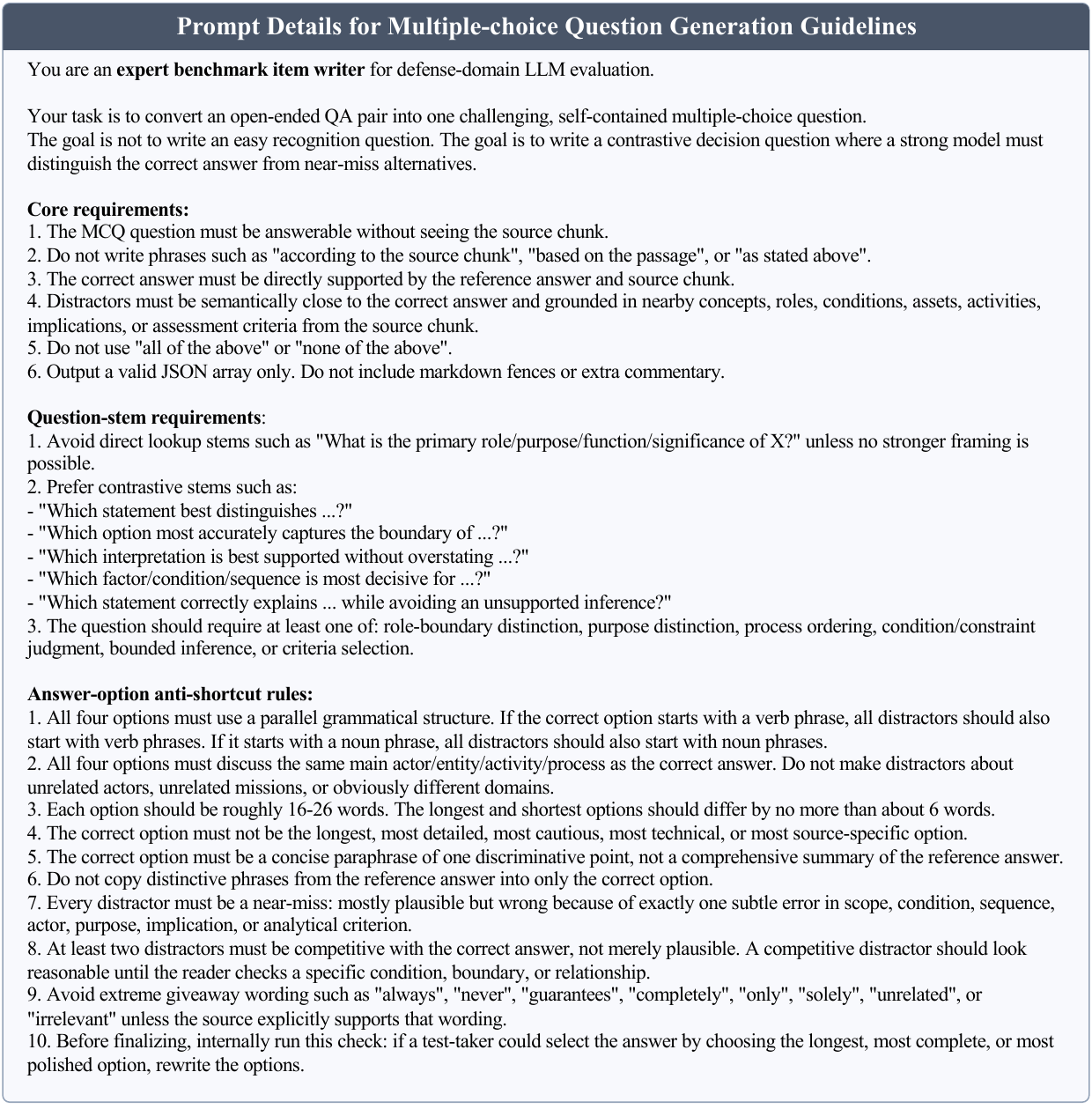}
  \caption{Prompt details for the multiple-choice question generation guidelines. The generator converts each open-ended QA pair into a challenging self-contained Multiple-choice Question using contrastive stems, source-grounded answers, and plausible distractors.}
  \label{fig:mcq_generation_guidelines_prompt}
\vspace{-1.em}
\end{figure*}
\begin{figure*}[h]
  \includegraphics[width=\linewidth]{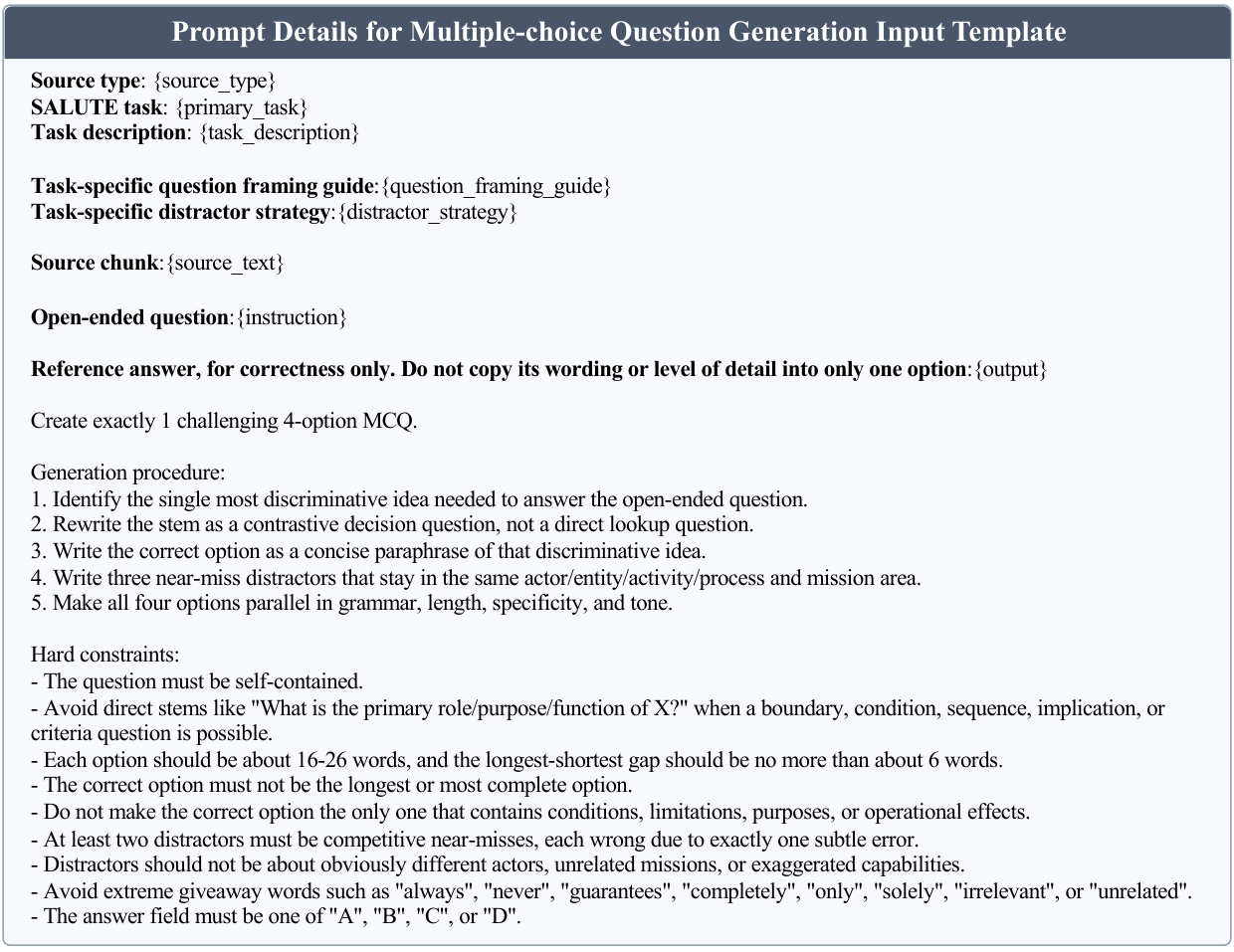}
  \caption{Prompt details for the multiple-choice question generation input template. The generator uses task metadata, source evidence, and an open-ended QA pair to create a self-contained Multiple-choice Question.}
  \label{fig:mcq_generation_input_template_prompt}
\vspace{-1.em}
\end{figure*}

For multiple-choice question (MCQ) generation, each open-ended QA instance is converted into a four-choice question grounded in the source chunk. 
From 8,163 open-ended QA instances, Qwen3-30B-A3B-Instruct-2507~\cite{yang2025qwen3} produces 8,139 valid MCQ candidates after excluding invalid outputs, forming the initial MCQ pool before subsequent filtering. 
We use the prompt in Figure~\ref{fig:mcq_generation_guidelines_prompt}--\ref{fig:mcq_generation_input_template_prompt}, which provides the source chunk, question, reference answer, and task information. 
The prompt instructs the model to avoid simple lookup questions and to write a self-contained stem. 
The correct option is concise and source-grounded, while distractors are close but incorrect near-misses.

\begin{figure*}[h]
  \includegraphics[width=\linewidth]{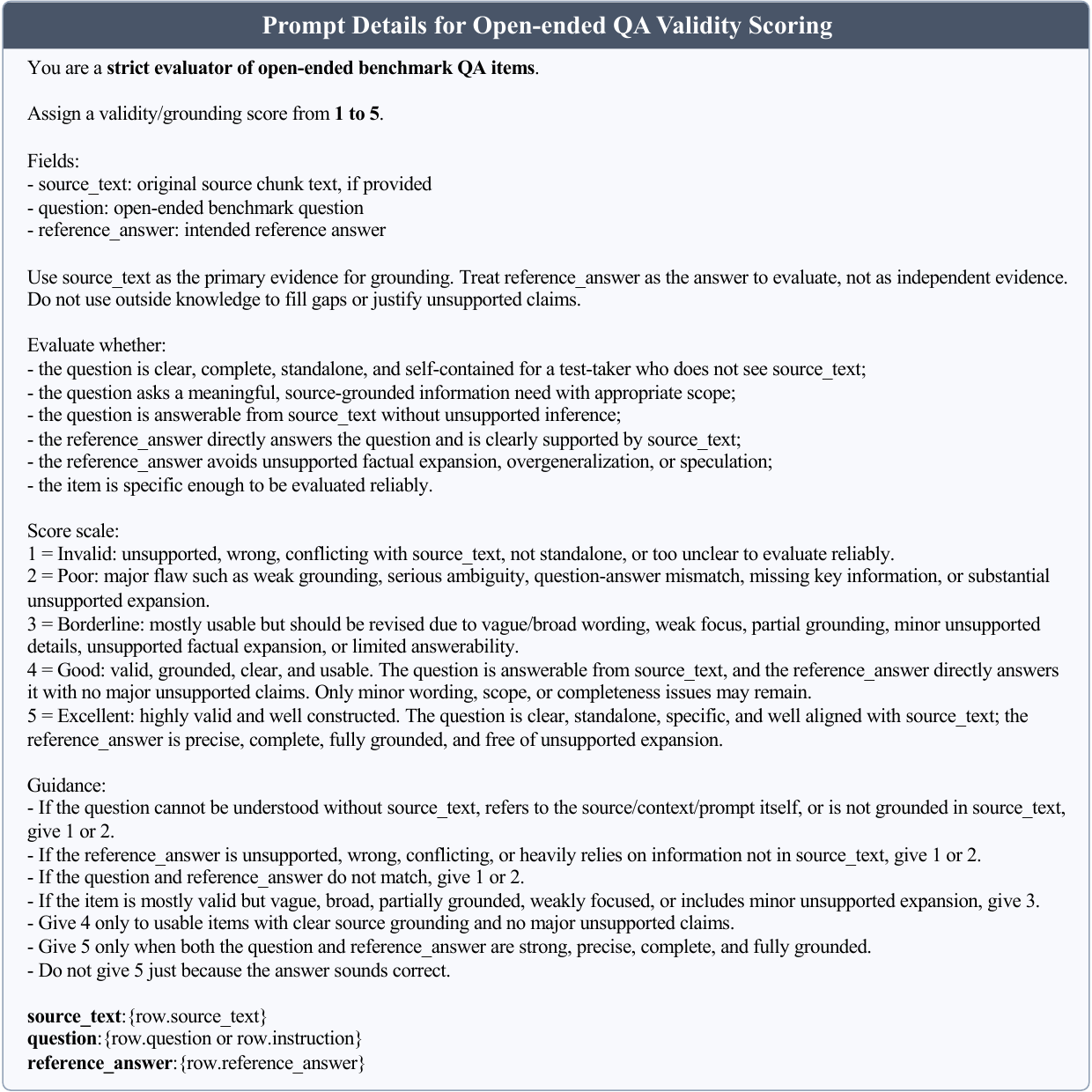}
  \caption{Prompt details for open-ended QA validity scoring. 
The evaluator scores each QA item for clarity, answerability, source grounding and reference-answer support using a 1 to 5 scale.}
  \label{fig:oe_qa_validity_scoring_prompt}
\vspace{-1.em}
\end{figure*}

\paragraph{GPT-based filtering.}
We apply GPT-5-based filtering to improve the quality of Salute-Bench candidates. 
For open-ended QA, GPT-5 assigns a 1--5 validity score based on question clarity, source-grounded answerability, and support for the reference answer, using the prompt in Figure~\ref{fig:oe_qa_validity_scoring_prompt}. 
Candidates scoring below 4 are discarded, and the resulting score distribution is shown in Figure~\ref{fig:filtering-oe-validity}.

\begin{figure*}[h]
  \includegraphics[width=\linewidth]{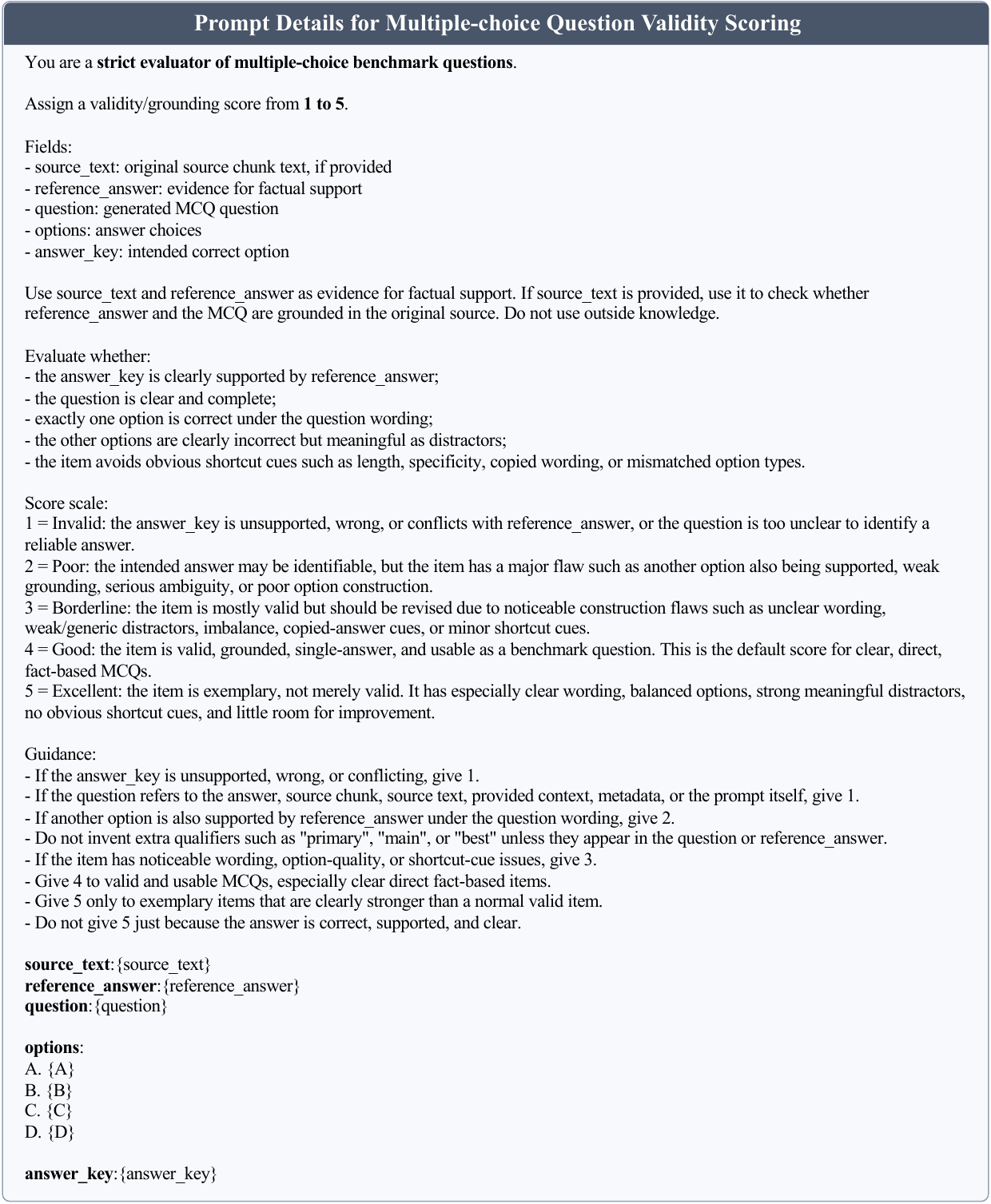}
  \caption{Prompt details for Multiple-choice Question validity scoring. 
The evaluator scores each Multiple-choice Question for answer-key support, single-answer validity, distractor quality and shortcut cues using a 1 to 5 scale.}
  \label{fig:mcq_validity_scoring_prompt}
\vspace{-1.em}
\end{figure*}

\begin{figure*}[h]
  \includegraphics[width=\linewidth]{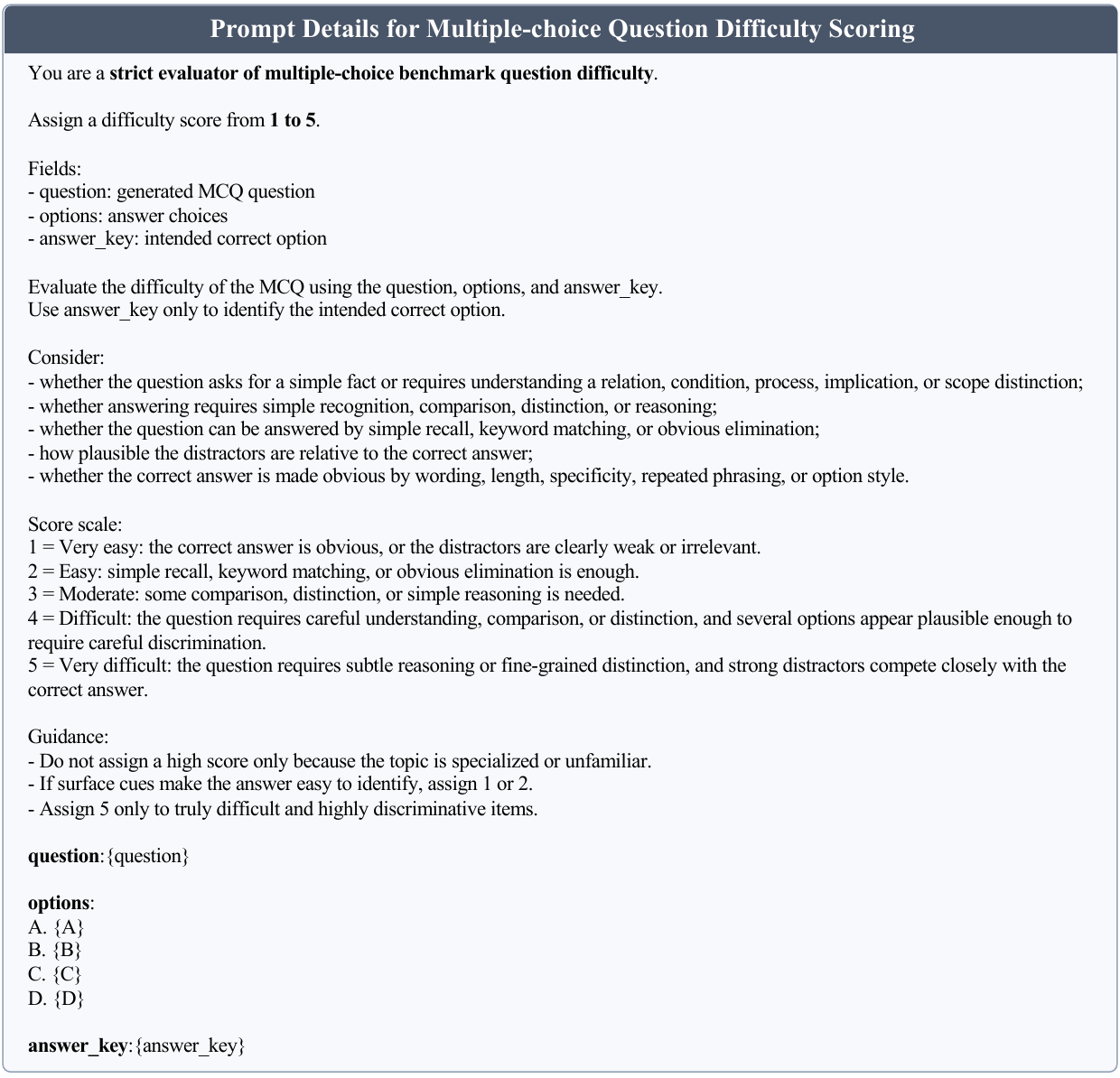}
  \caption{Prompt details for Multiple-choice Question difficulty scoring. The evaluator scores each Multiple-choice Question based on reasoning demand, distractor plausibility and shortcut cues using a 1 to 5 scale.}
  \label{fig:mcq_difficulty_scoring_prompt}
\vspace{-1.em}
\end{figure*}

For MCQ candidates, GPT-5 separately evaluates validity and difficulty. 
Using the prompt in Figure~\ref{fig:mcq_validity_scoring_prompt}, the validity evaluator measures source grounding, answer-key correctness, and single-answer consistency. 
The resulting validity score distribution is shown in Figure~\ref{fig:filtering-mcq-validity}. 
Using the prompt in Figure~\ref{fig:mcq_difficulty_scoring_prompt}, the difficulty evaluator assesses the need for reasoning beyond simple recall, the plausibility of distractors, and the absence of superficial cues that reveal the correct answer. 
The resulting difficulty score distribution is shown in Figure~\ref{fig:filtering-mcq-difficulty}. 
We retain only MCQ candidates whose validity and difficulty scores are both at least 4.

\begin{figure*}[h]
  \includegraphics[width=\linewidth]{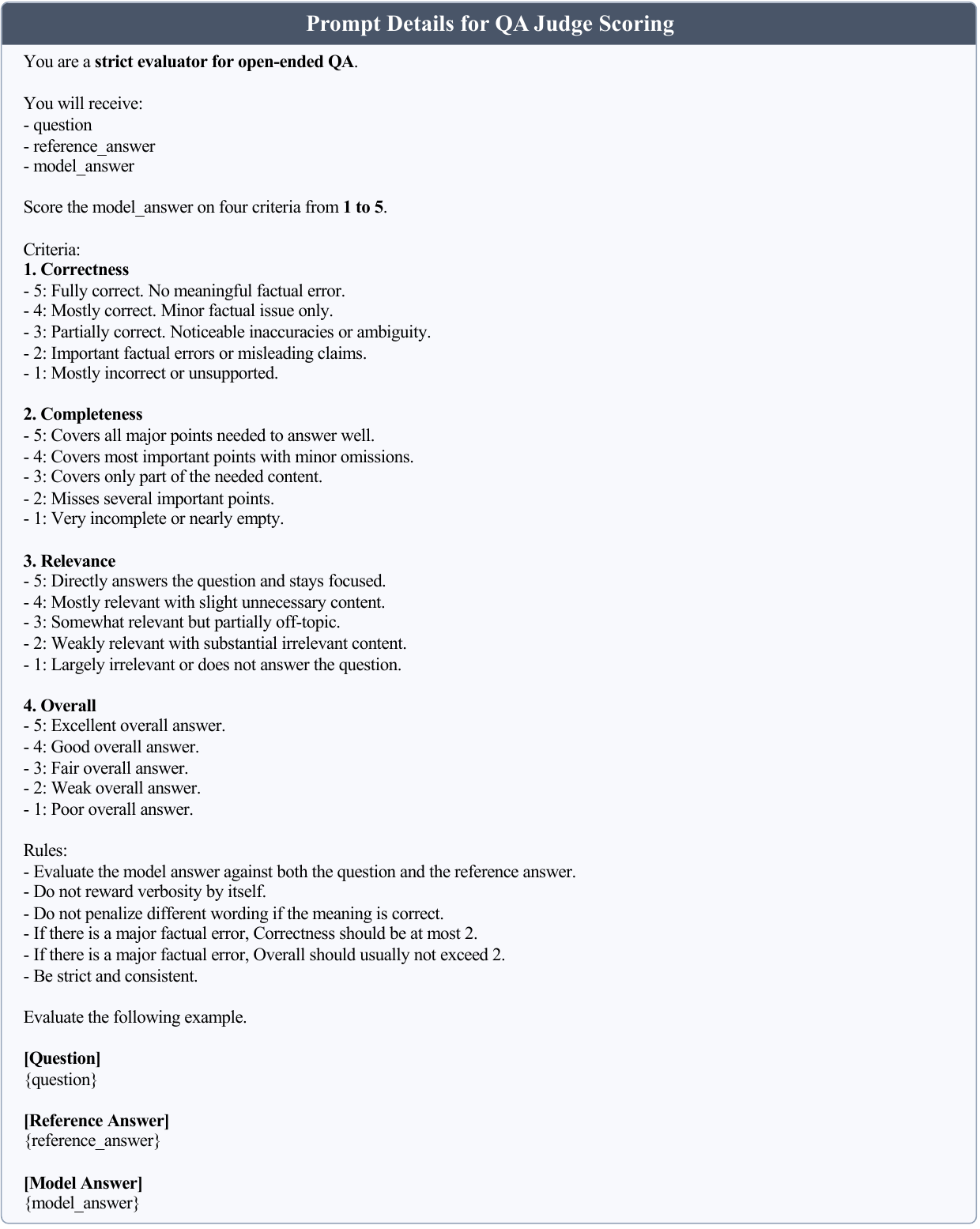}
  \caption{Prompt details for QA judge scoring. The judge scores each model answer for correctness, completeness, relevance and overall quality using a 1 to 5 scale.}
  \label{fig:qa_judge_scoring_prompt}
\vspace{-1.em}
\end{figure*}

\paragraph{Evaluation protocol.}
Salute-Bench consists of open-ended and multiple-choice questions constructed from held-out defense-domain source chunks. 
For open-ended questions, models generate free-form answers, which are evaluated using an LLM-as-judge protocol, a common scalable alternative to human evaluation for open-ended generation~\citep{liu2023g, zheng2023judging, kim2024prometheus}. 
To reduce ambiguity, the judge scores each response using a fixed rubric covering correctness, completeness, relevance, and overall quality. 
For multiple-choice questions, we evaluate accuracy by requiring models to select one of the given options. 
The full judge prompt and scoring rubric are provided in Figure~\ref{fig:qa_judge_scoring_prompt}.

\paragraph{Salute-Bench statistics.}
Table~\ref{tab:supp_salute_bench_stats} reports the length statistics of Salute-Bench. 
Open-ended questions have an average length of 27.44 tokens, with reference answers averaging 108.13 tokens, indicating that the benchmark requires explanatory defense-domain responses rather than short factual answers. 
Multiple-choice questions have a similar average length of 28.66 tokens, while each option averages 23.67 tokens. 
Together with our difficulty filtering, these statistics suggest that the multiple-choice split contains context-rich answer options and plausible distractors, helping reduce reliance on simple keyword matching.

\begin{table}[h]
\centering
\small
\begin{tabular}{llr}
\toprule
Split & Statistic & Value \\
\midrule
\multirow{2}{*}{Open-ended} 
& Avg. question length & 27.44 \\
& Avg. reference answer length & 108.13 \\
\midrule
\multirow{2}{*}{MCQ} 
& Avg. question length & 28.66 \\
& Avg. option length & 23.67 \\
\bottomrule
\end{tabular}
\caption{Length statistics of Salute-Bench. Lengths are computed using the Qwen3-8B tokenizer.}
\label{tab:supp_salute_bench_stats}
\vspace{-1.5em}
\end{table}

\subsection{Details of Defense Expert Assessment}
\label{sec:supp_expert_assessment}

We conduct an expert audit to further assess the quality of Salute-Bench after GPT-based filtering.
The audit is performed by five practitioners from a defense-domain company.

\begin{figure*}[!h]
  \includegraphics[width=\linewidth]{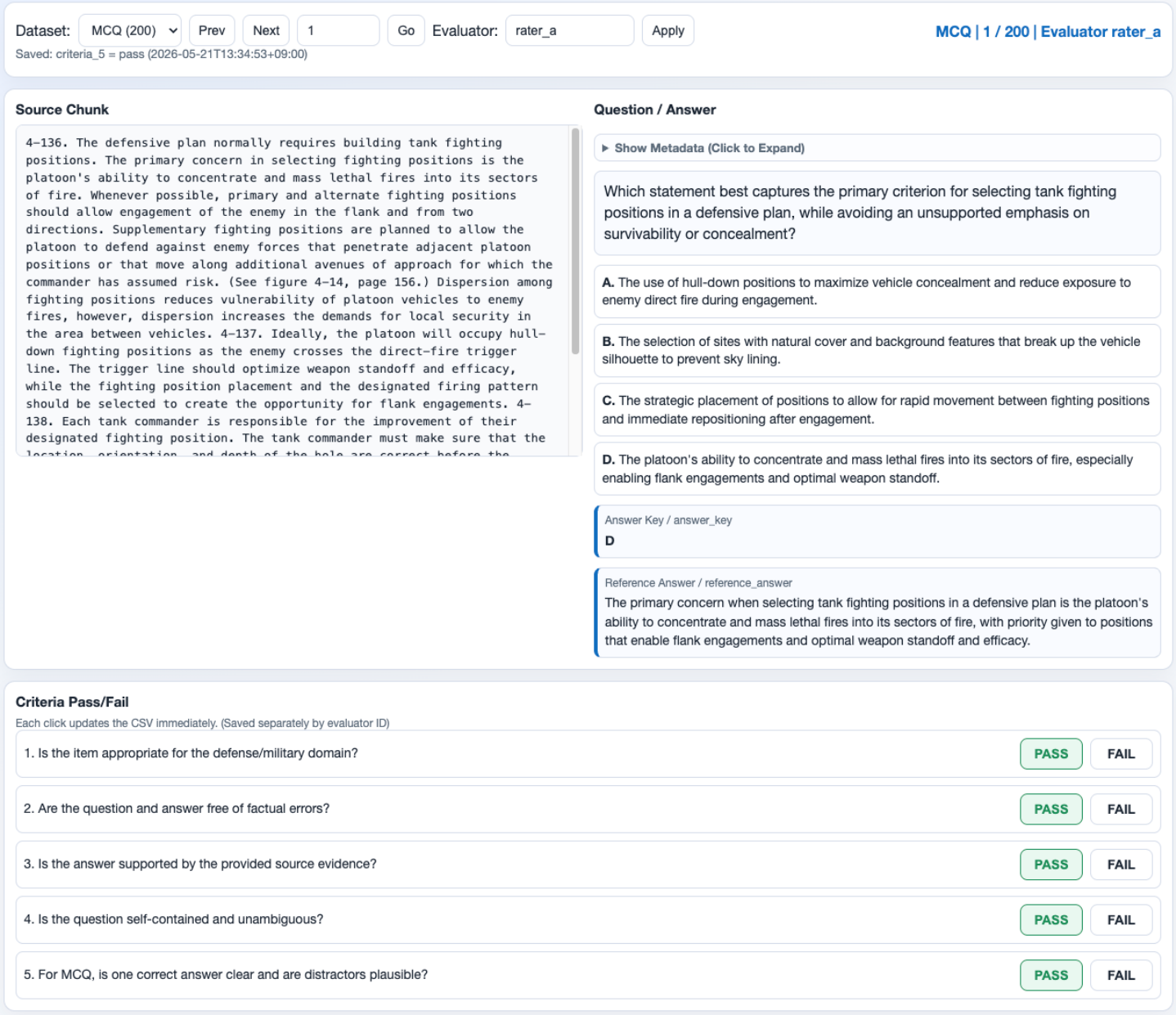}
  \caption{Human verification interface for Salute-Bench. 
  Annotators review each item with its source evidence and answer information, then mark pass or fail for domain appropriateness, factual correctness, source support, standalone clarity, and multiple-choice question option quality.}
  \label{fig:human_verification_interface}
\vspace{-1.em}
\end{figure*}

We randomly sample 400 final benchmark instances, including 200 open-ended questions and 200 multiple-choice questions.
For each instance, experts are provided with the source chunk used to construct the item, the question, and the reference answer.
For multiple-choice questions, experts are additionally provided with the answer options and the intended answer key.
As shown in Figure~\ref{fig:human_verification_interface}, experts assign a pass/fail judgment for each applicable criterion:

\begin{itemize}
    \item \textbf{Criteria 1:} Is the item appropriate for the defense/military domain?
    \item \textbf{Criteria 2:} Are the question and answer free of factual errors?
    \item \textbf{Criteria 3:} Is the answer supported by the provided source evidence?
    \item \textbf{Criteria 4:} Is the question self-contained and unambiguous?
    \item \textbf{Criteria 5:} For multiple-choice questions, is one correct answer clear and are the distractors plausible?
\end{itemize}

Criteria 1-4 are applied to all benchmark instances, while Criterion 5 is applied only to multiple-choice questions.
A practitioner accepts an instance if it fails at most one applicable criterion, and an instance is finally accepted only if all five practitioners accept it.
Under this rule, all 400 audited instances meet the acceptance standard.

\begin{figure}[t]
    \centering
    \includegraphics[width=\linewidth]{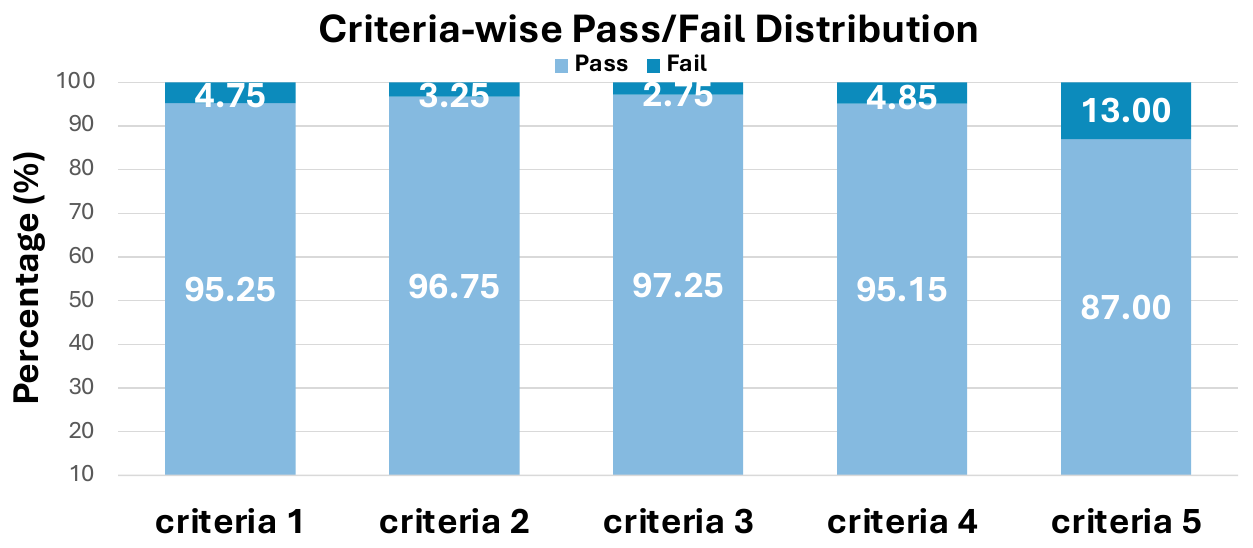}
    \caption{
    Criteria-wise pass/fail distribution from the defense expert audit.
    Criteria 1-4 apply to all audited instances, while Criteria 5 applies only to multiple-choice instances.
    Percentages are aggregated over expert-item judgments from five practitioners.
    }
    \label{fig:criteria_pf}
    \vspace{-1.5em}
\end{figure}

Figure~\ref{fig:criteria_pf} shows the criteria-wise pass/fail distribution aggregated over expert-item judgments.
All criteria achieve high pass rates, ranging from 87.00\% to 97.25\%, indicating that most audited items are appropriate, factually correct, source-supported, self-contained, and, for MCQs, single-answer valid with plausible distractors.
We further compute Fleiss' $\kappa$ for each applicable binary criterion across the five practitioners.
Across open-ended criteria 1--4 and multiple-choice criteria 1--5, Fleiss' $\kappa$ ranges from 0.800 to 0.983, with an average of 0.893, indicating strong inter-annotator agreement.
Overall, the expert audit provides an additional quality check beyond automated filtering and supports the reliability of Salute-Bench.

\section{More Experimental Results}
\label{sec:more_exp}

This section provides additional experimental analyses that complement the main results. We first examine the effect of replay data used during post-training and the impact of the CPT corpus filtering threshold. We then report task-wise and qualitative results, compare our adaptation pipeline with a retrieval-augmented generation baseline, and validate the reliability of the LLM-as-judge evaluation through human assessment.

\subsection{Replay Ablation}
\label{sec:supp_replay_ablation}
We analyze the effect of replay data at each post-training stage. 
For CPT, replay indicates the use of general-domain FineWeb~\cite{penedo2024fineweb} data mixed with Salute-Corpus.
For SFT, replay indicates the use of Tulu-3-SFT-Mixture~\cite{lambert2025tulu} instruction data mixed with Salute-Conv; the no-replay variant is trained only on Salute-Conv for two epochs.
For DPO, replay indicates the use of Salute-Pref as defense-domain preference replay together with Tulu-3-8B Preference Mixture~\cite{lambert2025tulu}.
Table~\ref{tab:supp_replay_ablation} reports Salute-Bench open-ended performance, Salute-Bench MCQ accuracy, and the average score across general LLM benchmarks.

\begin{table}[h]
\centering
\small
\setlength{\tabcolsep}{3.5pt}
\begin{tabular}{lcccc}
\toprule
Model & Replay & OE Avg. & MCQ & Gen. Avg. \\
\midrule
\multirow{2}{*}{Salute-Base}
& \xmark & 60.18 & \textbf{78.52} & 61.66 \\
& \cmark & \textbf{60.68} & \textbf{78.52} & \textbf{62.96} \\
\midrule
\multirow{2}{*}{Salute-Instruct}
& \xmark & 62.91 & 81.60 & 56.00 \\
& \cmark & \textbf{63.49} & \textbf{85.09} & \textbf{61.72} \\
\midrule
\multirow{2}{*}{Salute-LLM}
& \xmark & 63.89 & 84.06 & 64.68 \\
& \cmark & \textbf{64.51} & \textbf{86.02} & \textbf{64.69} \\
\bottomrule
\end{tabular}
\caption{Replay ablation results across post-training stages. OE Avg. denotes Salute-Bench open-ended average score, MCQ denotes multiple-choice accuracy, and Gen. Avg. denotes the average score across general LLM benchmarks.}
\vspace{-1.5em}
\label{tab:supp_replay_ablation}
\end{table}

Overall, replay data helps retain general capabilities while maintaining or improving Salute-Bench performance. 
This effect is most pronounced in SFT, where adding Tulu3 instruction replay raises the general benchmark average from 56.00 to 61.72, while also improving both Salute-Bench metrics. 
CPT replay improves open-ended and general performance without changing MCQ accuracy, and Salute-Pref replay further improves Salute-Bench performance with negligible change in the general benchmark average. 
These results indicate that replay data helps mitigate capability drift during defense-domain adaptation.

\subsection{CPT Filtering Threshold Ablation}
\label{sec:supp_filtering_ablation}

We analyze the effect of the corpus filtering threshold used during Salute-Corpus construction. 
Specifically, we train Salute-Base variants by removing the bottom 0\%, 15\%, and 30\% of chunks according to the ModernBERT-predicted quality scores. 
The 0\% setting uses the full deduplicated corpus without quality filtering, while the 30\% setting applies a more aggressive filter.

\begin{table}[h]
\vspace{-.5em}
\centering
\small
\setlength{\tabcolsep}{4pt}
\begin{tabular}{llccc}
\toprule
Model & Removed & OE Avg. & MCQ & Gen. Avg. \\
\midrule
\multirow{3}{*}{Salute-Base}
& 0\%  & 59.75 & 77.60 & 62.40 \\
& 15\% (ours) & \textbf{60.68} & \textbf{78.52} & \textbf{62.96} \\
& 30\% & 57.91 & 76.36 & 62.49 \\
\bottomrule
\end{tabular}
\caption{Salute-Corpus filtering threshold ablation. OE Avg. and MCQ denote Salute-Bench performance, and Gen. Avg. denotes the average score across general LLM benchmarks.}
\vspace{-1.0em}
\label{tab:supp_filtering_ablation}
\end{table}

Table~\ref{tab:supp_filtering_ablation} shows that removing the bottom 15\% of chunks yields the best overall performance across Salute-Bench and general benchmarks. 
Compared with no filtering, the 15\% threshold slightly improves both defense-domain performance and general benchmark average. 
In contrast, filtering 30\% degrades Salute-Bench performance, suggesting that aggressive filtering may remove useful domain-specific content.
These results support our choice of 15\% as a balanced filtering threshold.

\subsection{Task-wise Performance on Salute-Bench}
\label{sec:supp_taskwise}
Table~\ref{tab:task_wise_salute_bench} reports task-wise performance on Salute-Bench for Qwen3-8B, Qwen3-8B-Base, and Salute-LLM.
Salute-LLM outperforms both Qwen3 baselines across all eight tasks in both multiple-choice accuracy and open-ended QA average score.
These consistent task-wise gains indicate that the aggregate improvements are broadly distributed across doctrine- and news-oriented settings, rather than driven by a small subset of tasks.

\begin{table*}[t]
\centering
\small
\setlength{\tabcolsep}{5pt}
\renewcommand{\arraystretch}{0.95}
\begin{tabular}{lccc ccc}
\toprule
\multirow{2}{*}{\textbf{Task}} 
& \multicolumn{3}{c}{\textbf{Multiple-Choice QA Acc.}} 
& \multicolumn{3}{c}{\textbf{Open-Ended QA Avg.}} \\
\cmidrule(lr){2-4} \cmidrule(lr){5-7}
& \textbf{Qwen3-8B} 
& \textbf{Qwen3-8B-Base} 
& \textbf{Salute-LLM}
& \textbf{Qwen3-8B} 
& \textbf{Qwen3-8B-Base} 
& \textbf{Salute-LLM} \\
\midrule
\multicolumn{7}{l}{\textit{Doctrine Tasks}} \\
Conceptual Explanation      & 73.68 & 69.29 & \textbf{85.96} & 58.09 & 56.94 & \textbf{70.11} \\
Functional Role/Purpose     & 78.52 & 73.15 & \textbf{87.24} & 49.85 & 50.07 & \textbf{59.45} \\
Process/Structure           & 73.48 & 68.83 & \textbf{80.46} & 48.70 & 49.57 & \textbf{54.55} \\
Condition/Constraint        & 71.17 & 65.29 & \textbf{80.58} & 50.88 & 50.85 & \textbf{58.58} \\
\midrule
\multicolumn{7}{l}{\textit{News Tasks}} \\
Asset/Capability Role       & 74.28 & 75.71 & \textbf{82.85} & 58.96 & 59.30 & \textbf{67.29} \\
Activity/Purpose            & 87.69 & 81.53 & \textbf{98.46} & 62.05 & 65.09 & \textbf{71.56} \\
Implication/Impact          & 87.60 & 79.33 & \textbf{95.04} & 63.22 & 67.76 & \textbf{80.49} \\
Analytical Assessment       & 82.60 & 75.36 & \textbf{89.85} & 57.37 & 58.15 & \textbf{66.76} \\
\bottomrule
\end{tabular}
\caption{
Task-wise performance on Salute-Bench.
Multiple-Choice QA Acc. denotes multiple-choice accuracy, and Open-Ended QA Avg. denotes the average score across correctness, completeness, relevance, and overall quality.
}
\vspace{-1.0em}
\label{tab:task_wise_salute_bench}
\end{table*}

\subsection{Qualitative Results}
\label{sec:supp_qualitative}
\begin{figure}[t]
    \centering
    \includegraphics[width=\linewidth]{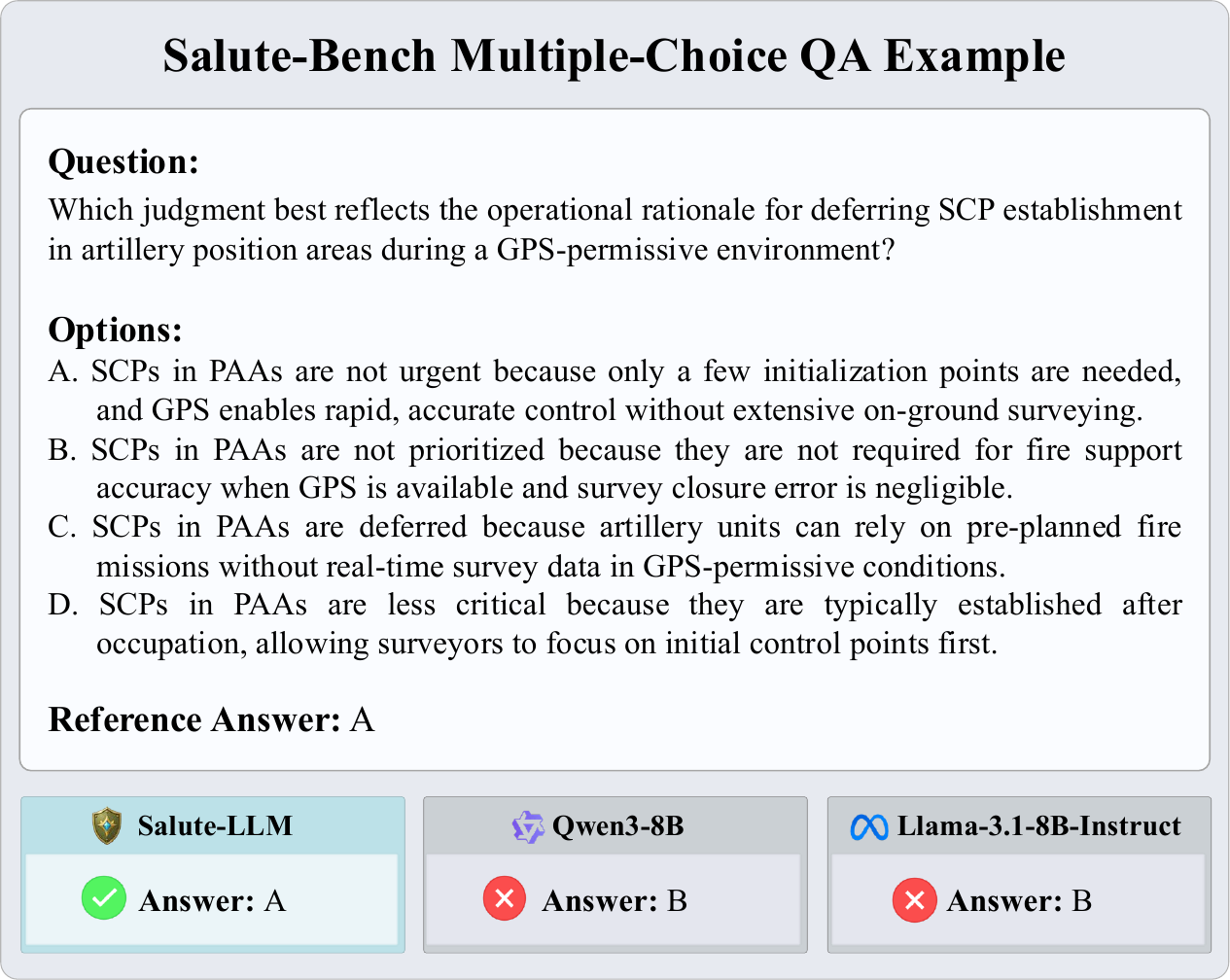}
    \caption{Qualitative multiple-choice example from Salute-Bench. 
Salute-LLM selects the correct operational rationale, while general-purpose baselines choose a plausible but incorrect distractor.}
    \label{fig:SALUTE_Bench_MCQ_example}
    \vspace{-1.5em}
\end{figure}

\paragraph{Comparison on Salute-Bench MCQ.}
Figure~\ref{fig:SALUTE_Bench_MCQ_example} presents a qualitative comparison on a Salute-Bench multiple-choice question. 
The question requires identifying the operational rationale for deferring survey control point establishment under GPS-permissive conditions. 
While the distractors contain related terminology about GPS, fire support accuracy, and survey procedures, only the correct option captures the key rationale that GPS reduces the urgency of establishing extensive on-ground survey control in artillery position areas. 
Salute-LLM selects the correct answer, whereas Qwen3-8B and Llama-3.1-8B-Instruct choose an overgeneralized distractor. 
This suggests that Salute-LLM better captures the doctrinal decision logic behind the scenario, rather than relying on surface-level associations with relevant military terms. 
This comparison further highlights the role of Salute-Bench in assessing defense-domain reasoning that requires distinguishing closely related but operationally different alternatives.

\begin{figure*}[t]
  \includegraphics[width=\linewidth]{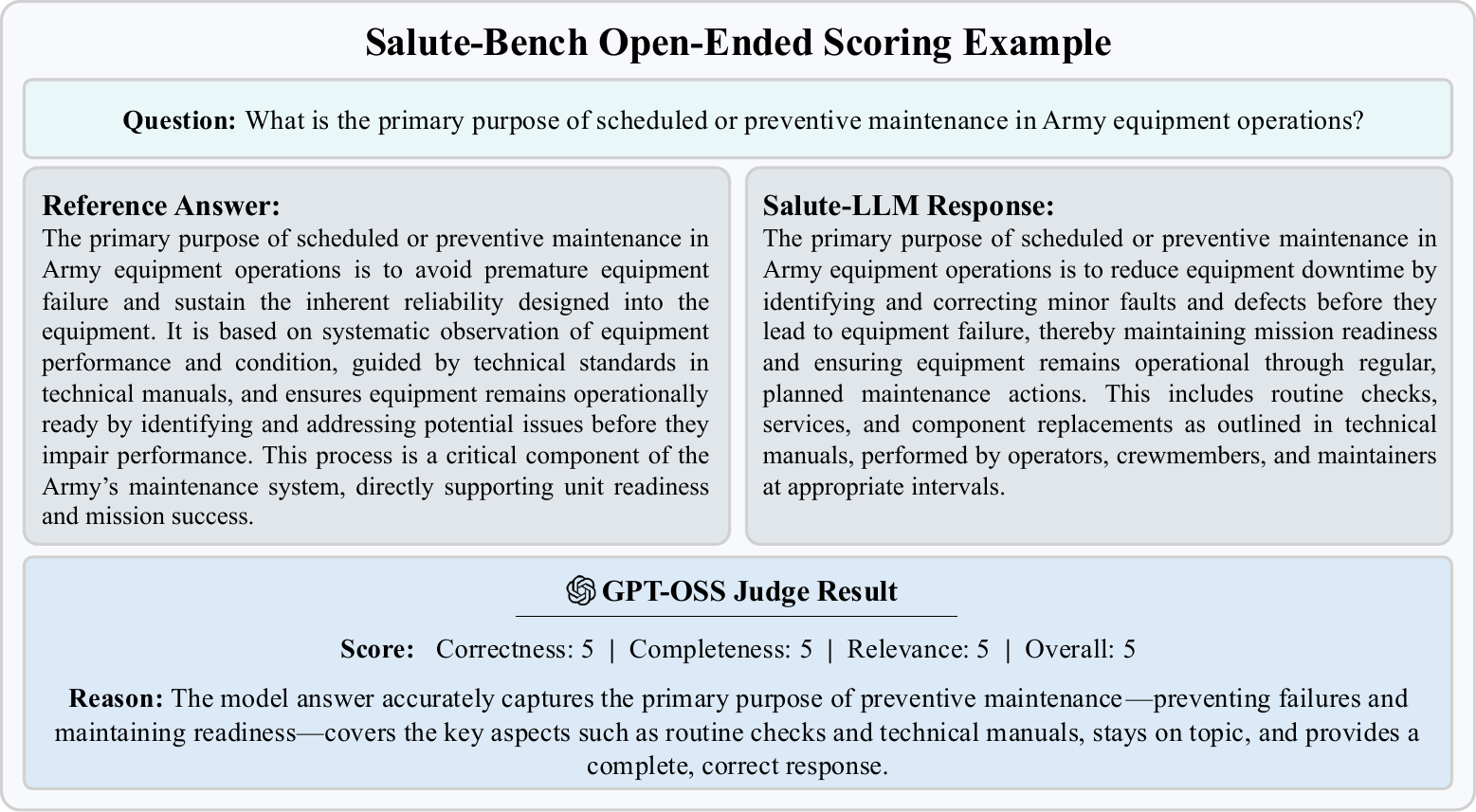}
      \caption{Qualitative example of GPT-OSS scoring on a Salute-Bench open-ended QA instance.
    The model answer receives full scores because it is correct, complete, and relevant with respect to the provided reference answer.
    }
  \label{fig:Judge_Result_Example}
\vspace{-.5em}
\end{figure*}

\paragraph{LLM-As-Judge Evaluation.}
Figure~\ref{fig:Judge_Result_Example} presents an example of the LLM-as-judge evaluation used for Salute-Bench open-ended QA. 
Given a question, reference answer, and model response, the judge assigns scores for correctness, completeness, relevance, and overall quality, together with a brief rationale. 
In this example, the model answer correctly identifies the purpose of scheduled or preventive maintenance as preventing equipment failure and maintaining readiness, while also covering routine checks and technical-manual guidance. 
The judge therefore assigns the highest score across all criteria. 
This example shows that the evaluation protocol rewards semantically aligned and complete answers rather than requiring exact lexical overlap with the reference.

\begin{table*}[t]
\centering
\scriptsize
\setlength{\tabcolsep}{10pt}
\begin{tabular}{lccccccc}
\hline
Model
& Correct.
& Complete.
& Relevance
& Overall
& OE Avg.
& MCQ
& Latency (s/query) $\downarrow$ \\
\hline
Qwen3-8B
& 54.70
& 46.95
& 69.34
& 49.85
& 55.21
& 77.28
& \textbf{0.37} \\

Qwen3-8B + RAG
& 58.62
& 55.93
& 70.63
& 56.37
& 60.39
& 80.88
& 3.94 \\

\textbf{Salute-LLM}
& \textbf{60.91}
& \textbf{59.87}
& \textbf{77.71}
& \textbf{59.53}
& \textbf{64.51}
& \textbf{86.02}
& \textbf{0.37} \\
\hline
\end{tabular}
\caption{
Comparison with a RAG baseline on Salute-Bench. OE Avg. and MCQ denote the open-ended average and multiple-choice accuracy, respectively, and latency is measured in seconds per query.
}
\vspace{-.5em}
\label{tab:rag_baseline}
\end{table*}

\begin{table*}[t]
\centering
\scriptsize
\renewcommand{\arraystretch}{1.15}
\setlength{\tabcolsep}{4pt}

\begin{tabular*}{\textwidth}{
    @{}l@{\extracolsep{\fill}}ccccccc@{}
}
\hline
&
\multicolumn{5}{c}{Human Evaluation Scores}
&
\multicolumn{2}{c}{Judge--Human Correlation}
\\
\cline{2-6}\cline{7-8}

Model
& Correctness
& Completeness
& Relevance
& Overall
& Average
& Spearman $\rho$
& Kendall $\tau$
\\
\hline

Qwen3-8B
& 49.6
& 43.2
& 80.2
& 48.4
& 55.4
& 0.816
& 0.726
\\

Qwen3-30B
& 52.6
& 48.6
& 85.0
& 51.6
& 59.5
& 0.783
& 0.688
\\

\textbf{Salute-LLM}
& \textbf{55.6}
& \textbf{51.8}
& \textbf{88.0}
& \textbf{55.6}
& \textbf{62.8}
& 0.811
& 0.713
\\
\hline
\end{tabular*}

\caption{
Human evaluation of model responses and per-model rank
correlations between human and GPT-OSS-120B scores.
Human evaluation scores are reported on a 0--100 scale.
}
\label{tab:human_evaluation}
\vspace{-1.em}
\end{table*}

\begin{figure*}[t]
  \includegraphics[width=\linewidth]{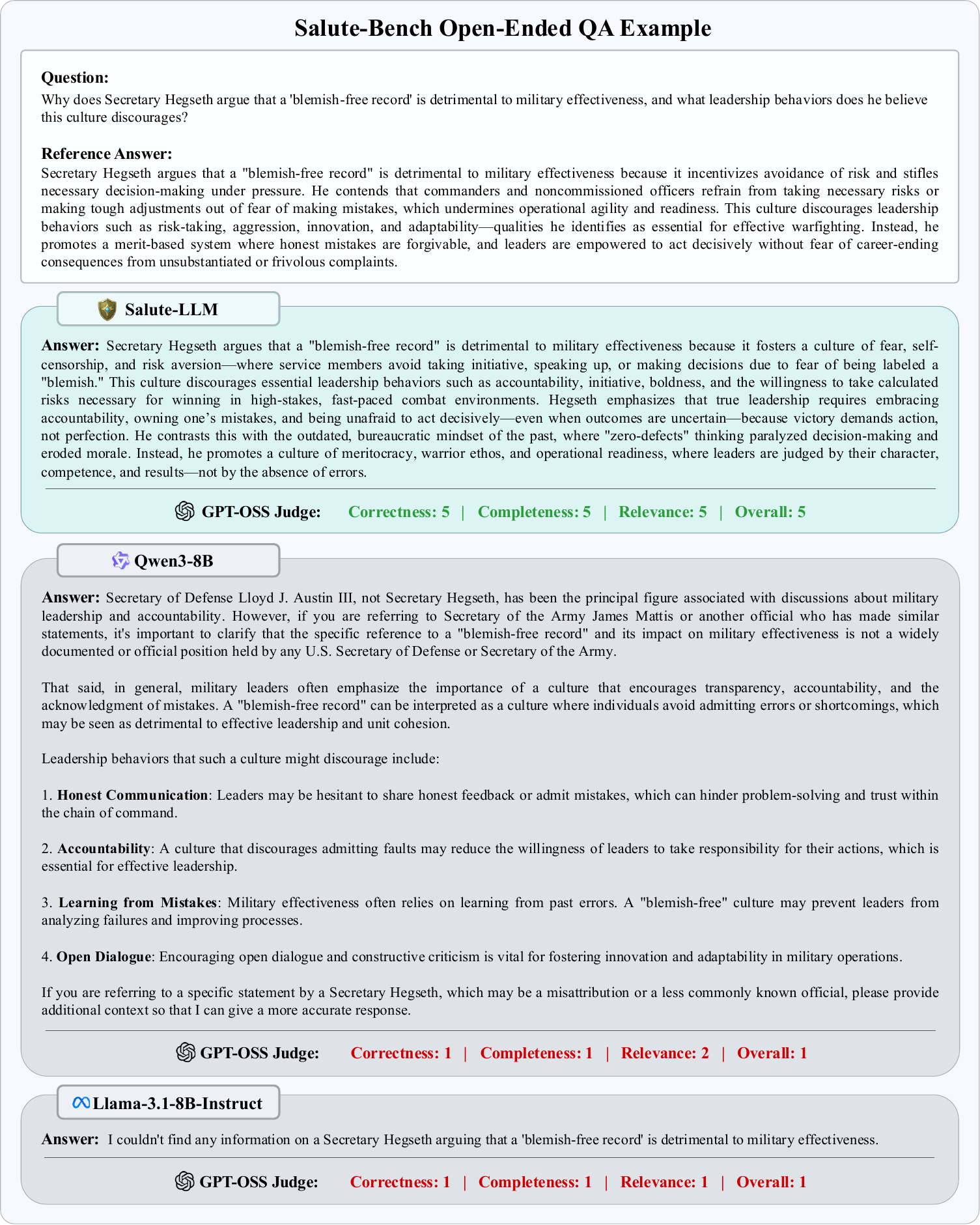}
  \caption{Qualitative open-ended QA comparison on Salute-Bench. Salute-LLM provides a reference-aligned defense-domain answer, while Qwen3-8B shifts to a generic discussion and Llama-3.1-8B-Instruct fails to engage with the question.}
  \label{fig:SALUTE_Bench_OE_example}
\end{figure*}

\paragraph{Comparison on Salute-Bench Open-ended QA.}
Figure~\ref{fig:SALUTE_Bench_OE_example} presents a qualitative comparison on a Salute-Bench open-ended question. 
The question asks why a ``blemish-free record'' may undermine military effectiveness. 
It also asks which leadership behaviors are discouraged under this culture.
Salute-LLM aligns closely with the reference by explaining that such a culture promotes fear and risk aversion, while discouraging initiative, boldness, decisive action, and calculated risk-taking.
As a result, it receives the highest judge scores across all criteria. 
In contrast, Qwen3-8B questions the premise and shifts to a generic discussion of leadership, while Llama-3.1-8B-Instruct abstains from answering. 
Their lower scores indicate that general-purpose models may over-generalize or fail to engage with defense-domain context, whereas Salute-LLM better captures the intended domain-specific reasoning.

\subsection{Retrieval-Augmented Generation Baseline}

We compare SALUTE's multi-stage adaptation with a no-training retrieval-augmented generation (RAG) baseline. Specifically, we augment Qwen3-8B with the same BM25/BGE-M3 hybrid retriever used in Section~\ref{sec:salute_conv}. To prevent benchmark leakage, the retrieval index excludes source chunks used to construct Salute-Bench. For each query, the top three retrieved chunks are provided to the model as supporting context.

As shown in Table~\ref{tab:rag_baseline}, RAG improves the open-ended average of Qwen3-8B from 55.21 to 60.39 and its multiple-choice accuracy from 77.28 to 80.88. Salute-LLM nevertheless outperforms the RAG baseline across all open-ended criteria, with gains of 4.12 points in the open-ended average and 5.14 points in multiple-choice accuracy. These results suggest that retrieved evidence alone does not fully substitute for defense-domain adaptation. Moreover, RAG increases inference latency from 0.37 to 3.94 seconds per query in our setup due to the additional retrieval step.

\subsection{Human Evaluation of Model Responses}

To assess the reliability of our LLM-as-judge evaluation, we conduct a human evaluation on 100 randomly sampled open-ended Salute-Bench instances. Two evaluators with defense-domain experience assess responses from Qwen3-8B, Qwen3-30B, and Salute-LLM using the same correctness, completeness, relevance, and overall criteria as GPT-OSS-120B. Human scores are averaged across the two evaluators and reported on the same 0-100 scale as the main evaluation.

As shown in Table~\ref{tab:human_evaluation}, Salute-LLM achieves the highest human-evaluated scores across all four criteria, consistent with the model ordering obtained from GPT-OSS-120B. We further compute the correlation between human and GPT-OSS-120B scores using the average of the four evaluation criteria for each response. Across models, Spearman correlations range from 0.783 to 0.816, while Kendall correlations range from 0.688 to 0.726. These results indicate strong consistency between human and LLM-based evaluations, suggesting that the reported open-ended gains are not solely an artifact of LLM judging.

\clearpage


\end{document}